\documentclass[12pt,a4paper,notitlepage]{report}
\usepackage{titlesec}
\titleformat{\chapter}[display]   
{\normalfont\huge\bfseries}{\chaptertitlename\ \thechapter}{20pt}{\Huge}   
\titlespacing*{\chapter}{0pt}{-50pt}{40pt}

\usepackage[left=3cm, right=3cm]{geometry}
\usepackage[affil-it]{authblk}

\usepackage[utf8]{inputenc}
\usepackage{amsmath}                            % (1)
\usepackage{amssymb}                            % (1)
\usepackage{amsthm}                             % (2)
\usepackage{graphicx}                           % (3)
\usepackage[pdfborder={0 0 0}]{hyperref}        % (4)
\usepackage{array}% http://ctan.org/pkg/array
\usepackage{booktabs}
\usepackage{multirow}
\usepackage{caption} 
\usepackage{amssymb}
\usepackage{multirow}
\usepackage{breqn}
\usepackage{enumitem}
\usepackage[usenames]{color}
\usepackage{pdflscape}
\usepackage{afterpage}
\usepackage{adjustbox}
\usepackage{longtable,pdflscape,booktabs}
\date{}
\def\L{{\cal L}}
\def\X{{\mathbf{X}}}
\def\x{{\mathbf{ x}}}
\def\w{{\mathbf{ w}}}
\def\a{{\mathbf{ a}}}
\def\R{{\cal R}}
\def\h{{\mathbf{ h}}}

\begin{document}

\title{Deep Learning Models for Digital Pathology}
\author{A{\"\i}cha BenTaieb and Ghassan Hamarneh}
\affil{Medical Image Analysis Lab, Simon Fraser University, Burnaby, BC, Canada}
% \institute{ School of Computing Science, Simon Fraser University,  Canada \\
% \{abentaie, hamarneh\}@sfu.ca}

\maketitle
\begin{abstract}

Histopathology images; microscopy images of stained tissue biopsies contain fundamental prognostic information that forms the foundation of pathological analysis and diagnostic medicine. However, diagnostics from histopathology images generally rely on a visual cognitive assessment of tissue slides which implies an inherent element of interpretation and hence subjectivity. Access to digitized histopathology images enabled the development of computational systems aiming at reducing manual intervention and automating parts of pathologists' workflow. Specifically, applications of deep learning to histopathology image analysis now offer opportunities for better quantitative modeling of disease appearance and hence possibly improved prediction of disease aggressiveness and patient outcome. However digitized histopathology tissue slides are unique in a variety of ways and come with their own set of computational challenges. In this survey, we summarize the different challenges facing computational systems for digital pathology and provide a review of state-of-the-art works that developed deep learning-based solutions for the predictive modeling of histopathology images from a detection, stain normalization, segmentation, and tissue classification perspective. We then discuss the challenges facing the validation and integration of such deep learning-based computational systems in clinical workflow and reflect on future opportunities for histopathology derived image measurements and better predictive modeling. 

\end{abstract}

% \keywords{digital pathology, histopathology, microscopy, deep learning, medical image analysis}

% \addtoToC{Table of Contents}%
\tableofcontents\vspace{-2ex}%
% \clearpage

% \addtoToC{List of Tables}%
% \listoftables%
% \clearpage

% \addtoToC{List of Figures}%
% \listoffigures%
% \clearpage

% \mainmatter%

\begin{table}[t] %% 
            \centering
            \caption{Nomenclature: list of used abbreviations and mathematical symbols. }
            \begin{adjustbox}{width=0.8\textwidth,totalheight=0.8\textheight,center=\textwidth,keepaspectratio}
            
            \begin{tabular}{ l l}
            \toprule
            Abbreviation & Description \\
            \midrule 
            WSI & Whole Slide Image \\
            CAD &Computer-aided diagnosis\\
            H\&E & Hematoxylin and Eosin \\
            IHC & Immunohistochemistry \\
            HPF & High Power Field \\
            IF & Immuno Fluorescence \\
            CD & Color Deconvolution \\
            SN & Stain Normalization\\
            LoG & Laplacian of Gaussian \\
            CNN & Convolution Neural Network \\
            FCN & Fully Convolutional Neural Network \\
            RNN & Recurrent Neural Network \\
            LSTM & Long Short Term Memory \\
            SAE & Stacked Auto Encoder \\
            SSAE & Sparse Stacked Auto Encoder\\
            VAE & Variational Auto Encoder \\
            GAN & Generative Adversarial Network\\
            cGAN & Conditional GAN \\
            F1 & F1 Score \\
            Acc & Accuracy \\
            P & Precision \\
            R  & Recall \\
            $N$    & Total dataset size \\
            $X$    & Input whole slide image \\
            $x$    & Input patch \\
            $Y$    & Ground truth whole slide image label\\
            $y$    & Ground truth patch label \\
            $\L(.)$   & Loss function \\
            $f_l(.)$    & Neural network function at layer $l$\\
            $\mathcal{W}$ & Neural network parameters\\
            \bottomrule
            \end{tabular}
            \end{adjustbox}
            
            \label{table:datasets}
\end{table}

% PART I INTRODUCTION
\chapter{Introduction}

Histology is the microscopic inspection of plant or animal tissue. It is a critical component in diagnostic medicine and a tool for studying the pathogenesis and biology of processes such as cancer, cell duplication or embryogenesis.

The clinical management of many systemic diseases, including cancer, is informed by histopathological evaluation of biopsy tissues, wherein thin sections of a biopsy are processed to visualize tissue and cell morphologies for signs of disease. Digital pathology incorporates the acquisition, management, sharing and interpretation of pathology information; including slides and data, in a digital environment. Digital slides are created when glass slides are captured, with a scanning device, to provide a high-resolution digital image that can be viewed on a computer screen or mobile device. Over the past few years, researchers have been starting to apply the tools of deep learning to scanned digital slides (i.e., histopathology images or whole slide images)  in order to perform tasks such as primary clinical diagnosis, secondary consultation, clinical outcome and analysis of abnormalities in tissues.

This first chapter gives a brief outline of the unique opportunities and challenges for deep learning applications in digital pathology then summarizes the purpose, scope and layout of this report.

\section{Digitized Pathology and Deep Learning: Challenges and Opportunities}

Over the last decade, the advent and subsequent proliferation of whole slide digital scanners has resulted in a substantial amount of clinical and research interest in digital pathology; the process of digitization of tissue slides.  Diagnosis from histopathology slides has many advantages: 1) fast acquisition (5-10 min diagnosis), 2) not as costly as other tests such as molecular profiling, 3) can be performed in real time or during surgery for guiding surgeons with frozen tissue sections, for instance, and  4) essential for diagnosis, oncology and personalized treatment. 

At the moment, a major limitation of the digital pathology slide is the unassisted human interpretation currently used for analysis. To promote consistency and objective inter-observer agreement, most pathologists are trained to follow simple algorithmic decision rules that sufficiently stratify patients into reproducible groups based on tumor type and aggressiveness. For example, in the most common group of brain tumors known as diffuse gliomas, the pathologist first begins by examining nuclear morphology to decipher a cell of origin (e.g., astrocytoma vs. oligodendroglioma). Once this first decision is established, the pathologist next assigns a degree of malignancy based on the presence of mitotic activity, tumor necrosis, and vascular proliferation~\cite{djuric2017precision}. Even with these simplified algorithms that focus on binary and sufficiently different features, inter-observer discordance still persist, even among sub-specialists~\cite{van2010interobserver}. This diagnostic uncertainty has promoted liberal and widespread use of costly molecular testing to differentiate between seemingly histologically indistinguishable lesions. Similarly, in efforts to maintain diagnostic objectivity, other potential prognostic and therapeutic morphologic biomarkers, such as foci of tumor-infiltrating lymphocytes and fibrotic tumor reaction, are often omitted. Indeed, even in the molecular era, the unassisted physician still largely relies on simple decision tree approaches that utilize only a small fraction of available molecular or genomics knowledge. This simplified approach to analyze histopathology is thus not fully leveraging the complex morphological information present for optimal patient treatment and outcome.

Interobserver agreement low scores encourage the introduction of computer-aided diagnostic (CAD) systems that can alleviate the need for human based decision rules and introduce more robust and accessible quantification systems. Specifically, deep neural networks have become the state-of-the-art machine learning based approach for most computer vision and medical image analysis tasks~\cite{litjens2017survey}. However, adopting deep models for digital pathology applications faces several challenges. Most of these challenges are related to the multi-magnification, high dimensional (i.e. millions of pixels) nature of histopathology slides and their acquisition.

\section{Purpose and Scope of this Survey}

In this report, we identified and reviewed 85 published works that form the state-of-the-art in terms of image analysis and deep learning methods tailored primarily for digital pathology images. Specifically, we discuss how while early attempts at detection and segmentation of tissue biomarkers in digital pathology images were rooted in traditional computer vision methods, there has been an evolution in the approaches in order to address the specific challenges associated with image analysis and classification of whole slide pathology images. Additionally, we discuss how the state-of-the-art deep neural networks are designed for high dimensional multi-magnification histopathology images as well as emerging research areas that are not always related to automating clinical tasks but a byproduct of computerized systems. We conclude the report with a discussion of some of the regulatory and technical hurdles that need to be overcome prior to the wide-spread dissemination and adoption of deep learning based solutions in clinical practice.

The remainder of this report is organized as follows: First, we briefly introduce the unique steps of the acquisition procedure of digital pathology slides. Then we review the different existing applications of deep learning models for digital pathology image analysis. We continue by giving a detailed overview of state-of-the-art deep learning methods from the datasets and architectures used to the validation strategies employed. Finally, we conclude with a summary and discussion and provide a Table~\ref{table:sota} categorizing the different works reviewed in the report.

% PART II 
\chapter{Image Acquisition in Digital Pathology}

Whole slide images (WSI) are digital images derived from biological specimens. Utilizing high-throughput, automated digital pathology scanners, it is possible to digitize an entire glass-mounted tissue slide observed under bright-field or fluorescent conditions, at a magnification comparable to a microscope. A major advantage of the digitization of tissue slides is that it facilitates sharing microscopy images between remote locations (i.e., telepathology) but also enables the integration of automated image analysis tools into pathology workflows and assist experts in the interpretation and quantification of biomarker expression within tissue sections~\cite{mccann2015automated}.   

\begin{figure} 
\centering{\includegraphics[trim=0cm 5cm 0cm 4cm, clip,width=\textwidth]{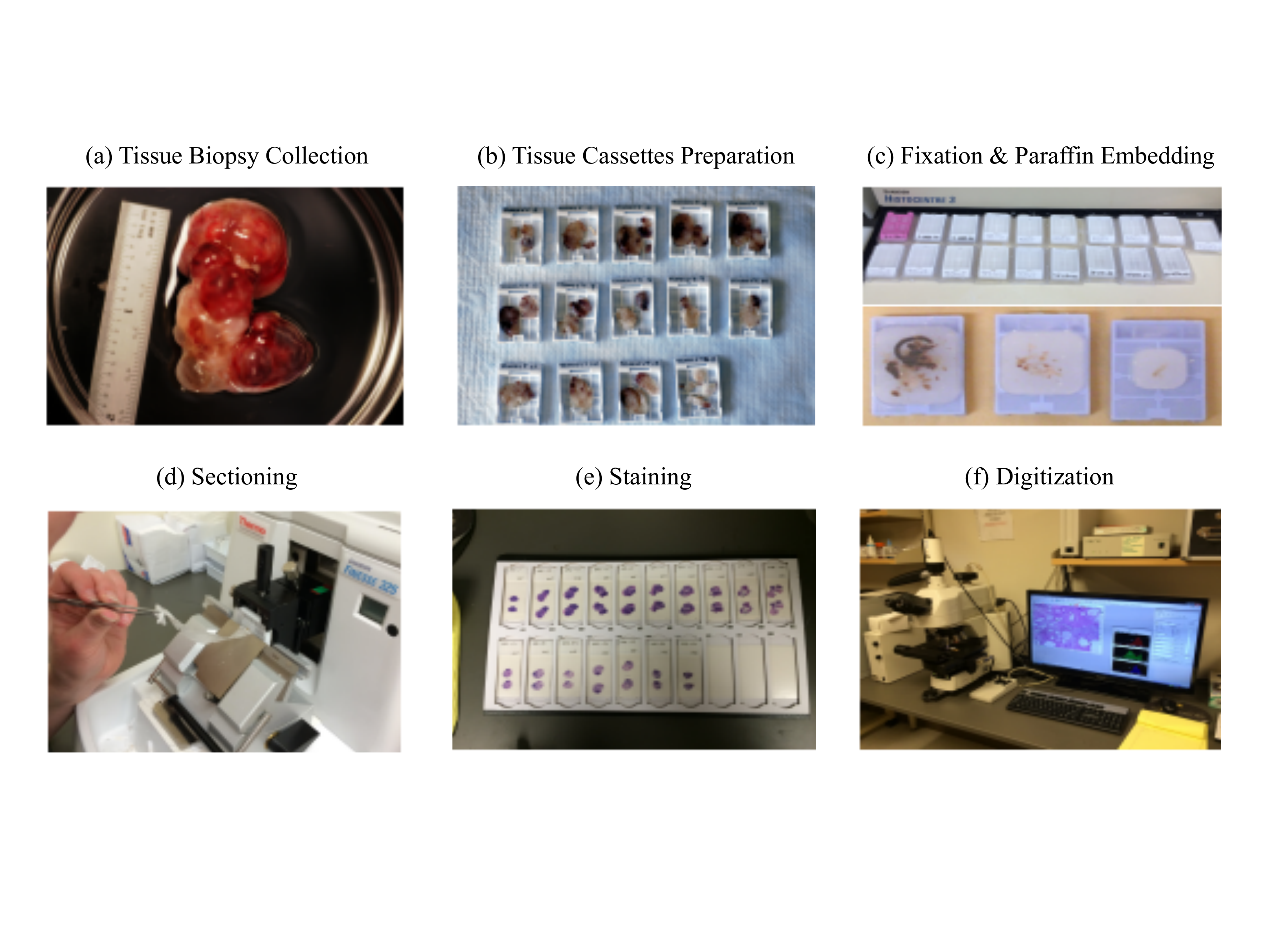}}
  \caption{Steps involved in the preparation of digital pathology images. Images are borrowed from~\cite{mccann2015automated}. }
   \label{fig:digit_steps}
\end{figure}

While digitizing pathology slides offers many advantages, the digitization procedure comes with important challenges that can hinder the visual analysis and computer-aided diagnosis of WSI. This chapter describes the step-by-step process of digitizing glass-mounted tissue slides and the challenges that derive from this procedure. Figure~\ref{fig:digit_steps} summarizes these steps.

\section{Tissue Preparation}

The process of preparing tissue slides for digital pathology analysis begins with a physician requesting a histology confirmation after assessing a patient with a physical exam and/or radiology and laboratory results. The clinical histology process starts by collecting enough good-quality tissue for a diagnosis. There are several approaches for tissue collection, including fine-needle aspiration, needle biopsy, excision biopsy or excision of a lesion in its entirety. Each of these collection techniques can alter the visual analysis of tissues and influence the final diagnosis accuracy. In fact, larger biopsies preserve more cellular context and allow the pathologist to examine multiple slides from different areas of the sampled tissue. Hence, the diagnosis is generally more reliable and accurate when determined from tissues collected with entire lesion excision as opposed to fine-needle aspiration. The sensitivity and specificity of computerized systems to the collection method are not clearly defined yet but one can assume that larger biopsy specimens would allow for a better diagnosis as they would result in more information (more images) but also more preserved context. After biopsy, a pathologist analyses the tissue at a macroscopic scale, measuring it and trimming it to fit into a tissue cassette container of size $10\times10\times3$ mm for the subsequent processing steps. 

At this stage tissues are transparent, soft and thick and will undergo series of steps allowing for their microscopy visual assessment. First, the collected tissue is immersed into a fixative solution that is used to stop cells from breaking down and tissues to be altered by microorganisms growth. After fixation, most tissues are embedded in a hardening material (i.e., paraffin waxes) to facilitate their sectioning (i.e., cutting) using a microtome. Certain tissues will require quick analysis and are frozen and sectioned in a cryostat (a microtome inside a freezer). Fixation and sectioning are critical steps in the preparation of histopathology slides as they prevent autolysis (i.e., cellular self-destruction by enzymes), allow the tissue to be kept close to its living state without loss of arrangement, and minimize changes in shape or volume often caused by subsequent steps. 

Once fixed and sectioned, tissue slices are still nearly invisible under a microscope as biological tissues are transparent. Therefore, another important step in the tissue preparation is tissue staining, which is the process of using dyes (generally chemical agents or antibody agents in the case of immunohistochemical staining) that have affinity for certain cell and extra-cellular components in order to create contrast. The chemical properties of these dyes produce the visual appearance that is seen under the microscope. In both diagnostic and research histopathology, the gold standard tissue sections are stained with hematoxylin and eosin (H\&E) where the cell nuclei are stained in blue by the hematoxylin which stains nucleic acids while cell cytoplasm is stained in pink/red by the eosin which stains proteins~\cite{buesa2007histology}. Finally, after staining, a coverslip (i.e., small glass sheet) is placed over the tissue mounted on the glass slide. This step creates an even thickness for viewing the tissue under a microscope and prevents the microscope lens from touching the tissue.

While some parts of the tissue preparation procedure can be automated in certain pathology centers (e.g., tissue fixation and sectioning can be done automatically with specific laboratory workstations), tissue handling involves manual intervention and introduces recognizable and well-documented artefacts in the case of paraffin fixed and H\&E stained tissue sections but less identifiable artefacts for other dyes. On average, tissue preparation takes 9 to 12 hours~\cite{litjens2017survey}. After preparation, tissues can be digitized.

\section{Tissue Slide Digitization} \label{sec:WSI}

Whole slide scanners are optical microscopes under robotic and computer control. These microscopes are mounted with highly specialized cameras containing advanced optical sensors that offer spatial resolutions of approximately 0.23-0.25 $\mu$/pixel using the 40x microscope objective~\cite{barisoni2017digital}. The essential components of a whole slide scanner, generally include the following: 1) a microscope with lens objectives, 2) light source (bright field and/or fluorescent), 3) robotics to load and move tissue slides around, 4) one or more digital cameras for capturing images, 5) a computer, and 6) software to manipulate, manage, and view the digitized slides.

Most whole slide scanners use a tiling or a line-scanning system to produce a WSI. Tiling consists of acquiring multiple individual high-resolution images as tiles while line-scanning creates linear scans of tissue areas. Both systems require stitching and smoothing the tiles or line scans together to create a single digital image of the histologic section. 

An important distinction between digital pathology and light microscopy resides in the concepts of magnification and resolution. In digital pathology, these concepts must be considered in the context of how images are acquired and displayed. In fact, magnification in light microscopy is determined by multiplying the power of the objective (4x, 10x, 20x or 40x) by the power of the eyepiece (generally 10x). This concept is not applicable in whole slide imaging as images are viewed on variably sized screens that can further amplify or shrink the original magnification. Hence, in digital pathology, for a whole slide image, the resolution is defined by the objective used to scan the slide (usually referred to as the WSI magnification) and the imaging sensor and is measured in micrometers per pixel. 

Once digitized, histopathology tissue sections are stored as WSIs which are digital files with varying sizes. The size of the file depends on the scanning objective and tissue size but commonly ranges from 200 MB to 10GB~\cite{webster2014whole}. In the context of health-care facilities, WSI files are significantly larger than digital image files routinely used in other clinical specialties such as radiology~\cite{farahani2015whole}. Therefore, compression-decompression methods, both lossy (e.g., JPEG2000) and loss-less (e.g., TIFF) types, are employed to store WSI. Currently, there is no standard file format for digital pathology images~\cite{webster2014whole}, although many vendors use the SVS format which stores a WSI as a multi-layered pyramid of thousands of image files with conserved filed of view and tile size spanning multiple folders. The pyramid representation enable the optimized real-time viewing of a WSI across multiple resolutions.  Although, the JPEG2000 compressed format is being used by some vendors there is an interest in migrating to the DICOM format as used in digital radiography~\cite{webster2014whole}.

\section{Artefacts} \label{sec:artefacts}

In pathology, artefacts are the result of the alteration of a tissue from its living state but in digital pathology, artefacts also include alterations of the rendered tissue image. 

The preparation of tissue slides and their digitization inevitably results in artefacts of various types that can compromise the image analysis and diagnosis. In fact, tissue appearance can be altered by the fixation, the specimen orientation in the block, the sectioning and the staining or immunolabelling steps which heavily dependent on human skills or, if automated, on human monitoring, machine maintenance and solution preparation~\cite{pichat2018survey}. Differences in protocols between pathology labs can greatly alter the appearance of even biologically similar tissue samples. Moreover, the digitization step can also introduce additional artefacts. In fact, depending on the digital scanner, the quality and resolution of the digitized tissue slide can vary significantly. 

A challenge in digital pathology is in identifying these artefacts and not confusing them with normal tissue components or pathological changes. In practice, pathologists learn to spot artefacts and depending on the extent of the damage on tissues, some sections often have to be manually discarded. Automatic systems can easily be sensitive to image and tissue artefacts which can cause the automatic image analysis systems to fail. For this reason, different pre-processing methods are generally used and most automatic systems have to account for inter and intra-slide variability. While designing automatic WSI analysis systems that are robust to common artefacts observed in digital pathology is still an open problem, feature learning via deep learning models enabled great progress in this direction. We discuss this further in the next chapter.

%========================================================================
% PART III
\chapter{Deep Learning Applications in Digital Pathology}

Recent applications of deep learning models in various fields resulted in redefining the state-of-the-art results achieved by earlier machine learning techniques. In digital pathology, deep neural networks, when appropriately trained, have proven capable of yielding diagnostic interpretations with accuracy similar to clinical experts~\cite{bejnordi2017diagnostic}. 

Prior to the success of deep learning models, traditional machine learning models already proved to be useful in digital pathology. In fact, unlike the simplified algorithms pathologists are trained to use,  machine learning models enable learning more complex decision functions. However, these conventional machine learning techniques usually do not directly deal with raw data but heavily rely on the data representations (i.e., hand-crafted features such as color, texture or shape), which require considerable domain expertise and sophisticated engineering. Many works have been proposed with hand-designed features that are often task specific, thus, do not generalize well across tissues and sites.  Deep learning models, by leveraging unsupervised or supervised feature learning, do not need heavy specialized applications, hence their quick success in digital pathology. In this chapter, we discuss the different applications of deep learning models for a variety of clinical and non-clinical tasks in digital pathology.

\section{Computer-aided Diagnosis: Automating Clinical Tasks}

 A number of image analysis tasks in digital pathology involve the quantification and highlight of morphological features (e.g., cell or mitotic count, nuclei grading, epithelial glands morphology). These tasks invariably require the identification (i.e., localization) of histologic primitives (e.g., cell, nuclei, mitosis, epithelium, cellular membranes, etc.). The presence, extent, size and shape or other morphological appearance of these structures are indicators of the presence or severity of disease. For instance, the size of epithelial glands in prostate cancer tend to reduce with higher Gleason patterns.  Another motivation for detecting and segmenting histologic primitives arises from the need for counting of objects, generally cells or nuclei. An example application is the Bloom Richardson grading system which is the most commonly used system for diagnosing invasive breast cancers and comprises three main components: tubule formation, nuclear pleomorphism, and mitotic count. Mitotic count, which refers to the number of dividing cells (i.e., mitoses) visible in H\&E tissue slides, is widely acknowledged as a good predictor of tumor aggressiveness. In practice, pathologists define mitotic count as the number of mitotic nuclei identified visually in a fixed number of high power fields (HPFs, 400x  magnification). Most of these tasks are time-consuming for the human-eye and can be highly sensitive to the level of expertise as well as the subjectivity of pathologists. 

The majority of deep learning models proposed for digital pathology use supervised learning in an attempt to automate the different parts of clinical experts' visual analysis tasks. Identifying histologic primitives is one area of applications for deep learning models. Another area is in the predictive modelling of outcome (i.e., cancer detection and classification, survival analysis) from WSIs which is the ultimate goal of the visual analysis of tissue slides. Both of these areas of application are motivated by the need for faster, more reliable and objective diagnostics and improved patients outcome.

\section{Non-clinical Tasks}

Another category of applications that leverage deep learning models is the result of computer-aided diagnosis itself. In fact, automatic systems generally require handling dataset-related challenges that are not necessarily critical to pathologists and are not directly involved in clinical tasks. Among such challenges, data harmonization, specifically stain normalization is an important factor that was shown critical in the development of computerized systems. While pathologists are less sensitive to tissue variability induced by staining inconsistencies, automatic systems often fail at generalizing to unseen datasets acquired with different staining protocols and can be highly sensitive to staining variations across tissue slides. Recently, deep models were proposed for staining normalization but also domain adaptation to handle such challenges. 

Another application concerns the generation of synthetic images via means of digital staining (transferring stains across images), virtual staining (automatically staining unstained tissues) or generative modelling (creating new images with realistic textures and stains).  There are different applications for these techniques that we discuss in the remainder of this report. One direct application of generative learning is data augmentation as a way to overcome the scarcity of available annotated datasets. 

Finally, a recent and important application of deep models to digital pathology concerns the design of interpretable computational systems. Interpretability is generally a desirable property for most computer-aided diagnostic systems as it facilitates clinical integration. However, existing systems are not generally optimized or designed for promoting interpretability of their outputs and underlying decision rules. Recently, there have been more attempts at designing interpretable deep models. 
%In fact, while pathologists are able to contextualize tissues in 3D, such information is not available to the computer system. Some works have been proposed to address such challenge and leverage stenography coupled with deep learning models to reconstruct volumes from tissue sections. 

%==================================================================================
% PART IV
\chapter{Deep Learning for Analyzing Digital Pathology Images}

A wide variety of deep learning models have been proposed for analyzing digitized tissue slides. Given the unique characteristics of WSIs many approaches present ways to process these high dimensional images and leverage their special multi-magnification nature.  
In this chapter, we discuss strategies to handle data scarcity and class imbalance. Then, we present the most commonly employed neural network architectures and describe the variety of deep learning systems, training strategies and validation procedures that have been proposed for use on WSIs.

\section{Digital Slide Representation and Datasets}\label{sec:data}

\begin{figure}  
\centering{\includegraphics[trim=0cm 5cm 0cm 5cm, clip,width=\textwidth]{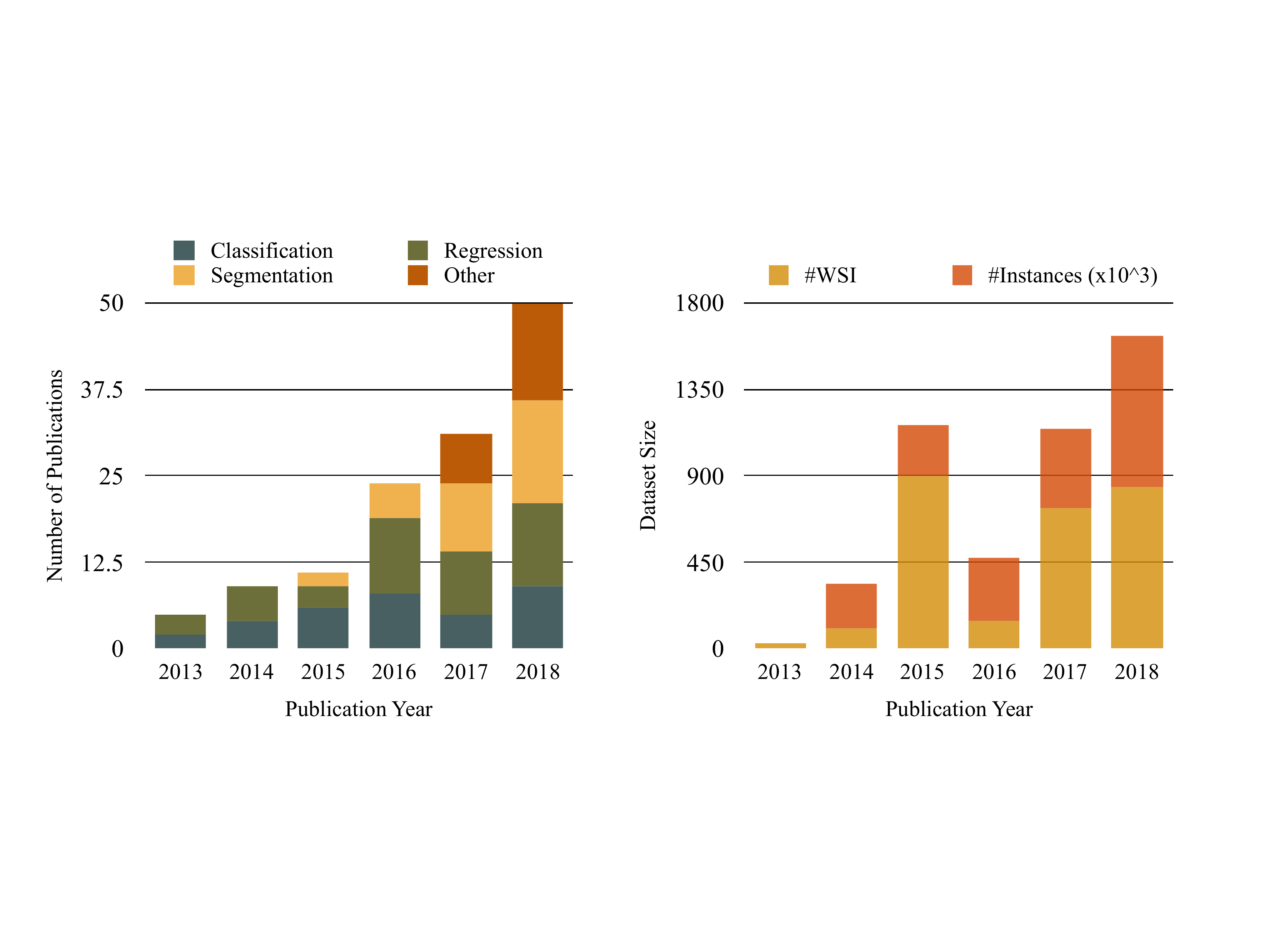}}
  \caption{Number of studies found per year categorized based on the type of task they address (left) and the average size of the dataset they use (right). The number of instances (\#instances) correspond to the average number of annotated instances available. }
\label{fig:stats}
\end{figure}

 Table~\ref{table:datasets} shows the different public datasets used in the studies we surveyed as well as their corresponding size. In the works we reviewed the datasets used covered a range of approximately 6 to 600 WSIs. While these are relatively small dataset sizes compared to what is usually necessary for training deep learning models (e.g., thousands or millions of images), a unique particularity of WSIs is their very large dimensions. 
As describe in section \ref{sec:WSI}, WSIs contain millions of pixels which can be leveraged efficiently to increase dataset sizes. Hence, all the works we surveyed in Table \ref{table:sota} used a variant of patch-based techniques to augment the dataset size and train deep learning models. For instance, for the task of detecting nuclei in tissue slides, the datasets created from sampled patches centered at annotated nuclei can contain thousands of positive instances from only dozens of available annotated WSIs. Figure~\ref{fig:stats} shows the sizes in terms of WSIs and their total corresponding annotated patches of the datasets used in different studies grouped per publication year. 

In practice, representing WSIs with patches is unavoidable. In fact, with current state-of-the-art computing resources, it is impossible to process a WSI in its entirety without extensively down-sampling the image which  would result in loosing most of the discriminative details and morphological features of the underlying tissues. 

Although patch-based representations are unavoidable, they do come with important shortcomings. First, patch-based representations imply the loss of global context captured within the multi-magnification levels of the tissue slide. In fact, tissues' structural organization is generally characterized by the different arrangements of cells and can only be observed at the lowest magnification levels (e.g., 20x or 10x) or with very large scale patches. Different works attempt to encode context when training deep models by using a pyramid representation in which input patches are extracted at different magnification levels (often $20x$ and $40x$) and processed with multi-scale~\cite{albarqouni2016aggnet} or cascaded~\cite{bejnordi2017context} deep network architectures.  

The second problem often faced when using patch-based representations is class imbalance. In fact, most applications involve datasets with very limited annotated positive samples. For instance, in the task of mitotic detection, only a few nuclei are generally labelled as mitotic, and the ratio of mitotic to non-mitotic nuclei can be up to 1 to 1000 which can arguable be defined as a highly imbalanced dataset. Class imbalance problems can be addressed by designing different patch sampling strategies that maximize the ratio of positives to negatives. One common approach is to densely sample positive patches and perform additional data augmentation with affine and elastic deformations on positive patches only. More sophisticated approaches involve crowdsourcing\cite{albarqouni2016aggnet} annotations, employing boosting techniques~\cite{cirecsan2013mitosis} or relying on active learning~\cite{yang2017suggestive}.

For some tasks such as cancer classification, it can be difficult to collect patch level annotations. In fact, most available dataset for cancer diagnosis are labelled at the slide-level only (i.e., an entire WSI is labelled as cancerous or not but the area of cancer is unknown). In these cases, it can be difficult to identify positive from negative patches. Most works treat patches as independent instances and extrapolate the slide-level label to all sampled patches from a given WSI. While this approach can be efficient, it is fundamentally flawed and results in high false positive patch predictions. A more accurate approach involves formulating classification and segmentation tasks with slide-level annotations only as weakly labelled problems and using machine learning frameworks such as multiple instance learning~\cite{jia2017constrained} to train the deep learning models.  Other strategies involve designing different aggregation techniques to infer a slide-level prediction from all processed patches at inference time. In this case, a trained model is applied on densely sampled patches from a test WSI in a sliding window fashion and a slide-level prediction score map is obtained. The most successful aggregation strategies involve training secondary machine learning models (e.g., random forest)~\cite{bejnordi2017diagnostic} as classifiers on hand-crafted features  (e.g., detected tumor size) extracted from the prediction score map obtained from the patch-level deep model.

Finally, a commonly used strategy for overcoming small dataset sizes (despite the large number of patches that can be extracted from a WSI) is to rely on pre-trained deep learning models. In such case, models are first trained on very large datasets from other domains (e.g., natural scene images) and fine-tuned on the smaller available digital pathology datasets. For instance, it has been shown that convolutional neural network (CNN) architectures trained on natural scene images can generalize relatively well (even without fine-tuning) to a variety of digital pathology tasks~\cite{bejnordi2017diagnostic, chen2017dcan, malon2013classification, wang2016deep, phan2016transfer}.

%===> Figure with size of datasets per year and per site 
%===> table listing all publicly available datasets 

\section{Deep Learning Architectures}

The majority of deep learning models used in digital pathology are of type feed forward or recurrent neural networks. Neural networks are commonly associated with acyclic graphical models that describe a composition of many different functions $f$ approximating some unknown function $f^\ast$. The goal of a neural network is to define a mapping $f(X, \mathcal{W})$ of an input $X$ and learn the parameters $\mathcal{W}$ that result in the best function approximation.  %Figure with architecture size per year 

\subsection{Convolutional Neural Networks}

In digital pathology, CNNs are the most commonly used type of feed forward neural networks. A CNN is a composition of a sequence of $L$ layers (i.e., functions) that maps an input image $\X$ to an output vector $Y$ (e.g., a scalar or output vector) such that: 

\begin{eqnarray}
    Y &=& f(\X; \w_1, \w_2, \dots, \w_L) \\
      &=& f_L(\a_{L-1}; \w_L) \circ f_{L-1}(\a_{L-2}; \w_{L-1}) \circ \dots \circ f_2(\a_1; \w_2) \circ f_1(\X; \w_1)
\end{eqnarray}
\noindent where $\mathcal{W} = \{\w_1, \w_2, \dots, \w_L\}$ are the trainable parameters (weights and biases) at different layers, $\{\a_1, \a_2, \dots, \a_L\}$ represent the intermediate outputs at each layer that form the networks' features or internal activations. Conventionally, the successive layers of a CNN model are defined to perform one of the following operations: i) 2D or 3D convolutions with trainable filter banks, ii) spatial pooling (e.g., average or max pooling), iii) non-linear activations (e.g., rectified linear units, hyperbolic tangent, sigmoid). Generally, the final layer of a CNN is represented as a fully connected or dense layer that maps the penultimate output activations to a distribution over categories $P(Y|\X; \mathcal{W})$ through a softmax function.   

Among the works we surveyed in Table \ref{table:sota}, the majority used CNN architectures with relatively shallow architectures (i.e., 2 to 8 layers) when it comes to tasks related to the analysis of histologic primitives (e.g., nuclei and cell localization, cell classification).  For instance, many works~\cite{gao2017hep, han2016hep, phan2016transfer, malon2013classification} adopted LeNet~\cite{lecun1998gradient} and AlexNet~\cite{krizhevsky2012imagenet} architectures with minor modifications to the penultimate layer output size.  Both of these networks are relatively shallow, consisting of two and five convolutional layers, respectively and employed convolutional kernels with large receptive fields in early layers and smaller kernels closer to the output. The main distinction of the AlexNet architecture is the use of rectified linear units instead of the hyperbolic tangent as activation function. 
In contrast, most prediction models we surveyed (e.g., cancer prediction and grading) rely on deeper network architectures (i.e., 8 to 150 layers)~\cite{bauer2016multi, liu2017detecting, bejnordi2017deep, wang2016deep}. Generally, these architectures are adapted from the VGG-16~\cite{simonyan2014very}, Inception~\cite{szegedy2016rethinking} and ResNet~\cite{he2016deep} models which are all built on small fixed-size kernels in each layer. These CNN architectures introduced novel building blocks that were shown to improve training efficiency and  reduce the total amount of trainable parameters. Specifically, the Inception model consists of inception blocks where instead of having a single convolution layer applied to a given input, the model uses multiple parallel operations (i.e., convolutions with varying kernel sizes and pooling) applied to the same input. This strategy was shown to act as a multi-level feature extraction scheme. In the ResNet architecture, residual blocks are introduced in order to train very deep models (i.e., 150 layers) effectively. Instead of learning a function, ResNet blocks only learn the residual and are preconditioned towards learning mappings that are close to the identity function. 
Finally, other works that did not use the above mentioned state-of-the-art architectures~\cite{cruz2014automatic, litjens2016deep, cruz2017accurate, yao2016imaging} designed custom CNN models where the building blocks are typically convolutions with varying kernel sizes depending on the dataset and task, max pooling layers, rectified linear units as activations and fully connected layers that act as final classifier. 
An example of a common custom architecture is the multi-scale CNN used for mitosis~\cite{albarqouni2016aggnet} and cell~\cite{song2015accurate} detection. Multi-scale architectures were proposed to incorporate larger context when using patch-based representations of WSIs. A multi-scale CNN is composed of multiple parallel trainable CNN architectures where each network includes a filter bank layer, non-linearity functions and pooling operations and is trained on similar size patches extracted at increasing input image scales. The output predictions obtained for all patches are aggregated using an additional fully connected layer or using a simple geometric average over all prediction scores to predict a single categorical output.

Another application of CNNs in digital pathology is as pixel-level classifiers in segmentation tasks. To predict a class label for each pixel in an input image, a common approach is to use a variant of a CNN named fully convolutional network (FCN)~\cite{long2015fully}. The particularity of FCNs is that they can  receive inputs of arbitrary size and produce correspondingly-sized outputs by removing all fully connected layers and introducing upsampling layers to supplement the usual contracting CNN network composed of stacked convolution, non-linearities and pooling (or downsampling) layers. There are different variants of FCN architectures but the most popular model used in digital pathology is the UNet~\cite{ronneberger2015u} which uses VGG-16 as a contracting network and combines it with its symmetric counterpart where all pooling layers are replaced by upsampling operations to increase the resolution of the output and form a u-shaped architecture. Skip connections are used to propagate information across the network and facilitate training.

 Whether they are used for classification, detection or segmentation, the parameters $\mathcal{W}$ of feed forward neural networks (i.e., CNNs and their variants) are generally trained in a supervised setting by optimizing a cost function. Given a set of $N$ annotated training instances $\{(\X^{(i)}, Y^{(i)})\}$, the parameters $\mathcal{W}$ can be estimated by solving the following optimization problem: 

\begin{equation}\label{eq:optim}
    \min\limits_{\mathcal{W}} \sum\limits_{i=1}^N \L(P(Y^{(i)}|\X^{(i)}; \mathcal{W}), Y^{(i)}) + \R(\mathcal{W}),
\end{equation}
\noindent where $\L$ is a defined cost function and $\R$ is a regularization term over the trainable parameters $\mathcal{W}$. 
In most cases, weight decay is used as regularization and the cost function is defined as the cross-entropy loss between the training data and the model's predictions: 

\begin{equation}\label{eq:CE}
    \L = - \log  \Big[P(Y^{(i)}|\X^{(i)}; \mathcal{W})\Big].
\end{equation}

Common variants of the cross entropy loss in digital pathology involve adding a weighting coefficient to handle class imbalance~\cite{xie2015deep, ronneberger2015u, sirinukunwattana2015spatially} or introducing additional auxiliary terms~\cite{chen2017dcan} in the loss defined in eq.(\ref{eq:CE}). 

Most deep models are optimized following eq.(\ref{eq:optim}) in an end-to-end fashion using stochastic gradient descent. In order to compute the gradients of the cost function  (eq.(\ref{eq:CE})) with respect to the network's parameters, the backpropagation algorithm is used to allow the information from the cost function to flow backward  in a recursive fashion.

\subsection{Recurrent Neural Networks}

Another category of neural networks often used in digital pathology are the recurrent neural networks (RNNs) which are generally used to model sequential data where the input and output (as in feed forward neural networks) can be of varying length. In its simplest form, the particularity of an RNN is in the capacity to maintain a latent or hidden state $\h$ at a given time $t$ that is the output of a non-linear mapping from its input $\X_t$ and given the previous state $\h_{t-1}$:

\begin{equation}
    \h_t = f(\h_{t-1}, \X_t; \mathcal{W}),
\end{equation}
where $\mathcal{W}$ are the trainable parameters and $f$ is the non-linear mapping function that is repeatedly applied to all input elements of the sequence $\X$. For classification, one or more fully connected layers
are typically added followed by a softmax to map the sequence to a posterior over the classes.

The training procedure for a RNN is similar to the one described above in eq.(\ref{eq:optim}) with the difference being that the total cost function for a given sequence $\X$ paired with a sequence $Y$ is simply the sum of the costs over all time steps (i.e., backpropagation through time). 

Different variants of RNNs have been proposed. In digital pathology, the most commonly used recurrent models are gated RNNs and long short term memory (LSTM) networks  that use additional gates (i.e.,  trainable non linear functions) to accumulate and forget information over a long duration of time. LSTMs, as opposed to gated RNN, incorporate gated self loops that are conditioned on the current state and context. These self loops are primarily introduced to facilitate training by modelling long-term dependencies and avoiding the gradients to vanish through the successive time steps. 

In the works we reviewed, RNNs and LSTMs have been primarily used for segmentation tasks. In these applications, 2D WSIs are modelled as a sequence of patches and dependencies between local patches are modelled using the recurrent architectures. In order to model dependencies in both axis of the image (i.e., rows and columns), most works use 2D versions of the recurrent networks. For instance, Xie et al.~\cite{xie2016spatial} proposed a 2D spatial RNN to segment muscle perimysium in digital pathology images where each input image is partitioned into non overlapping patches. In order to process an input image, the image patches are sorted in an acyclic sequence. Sub-hidden states are introduced and defined as a weighted combination of inputs from the 4-connected adjacent patches (left, right, top and bottom) within rows and columns of a grid representing the organization of all patches in the image. 
Similarly, 2D LSTMs can be used~\cite{agarwalla2017representation} to capture contextual information by processing WSIs as a 2D sequence of non-overlapping neighbouring patches. The input to the 2D LSTM model is represented as a sequence of two multi-dimensional vectors (e.g., feature vectors obtained with a CNN). Each 2D LSTM unit is pair-wise connected to its 4-connected neighbors and is composed of twice the number of gates as exist in 1D LSTM units (i.e., input, output, forget and cell memory gate) to process inputs from neighbouring patches along the columns and rows of the original WSI.  

Generally, recurrent models are integrated within CNN models to obtain feature representations of the input images and do not operate directly on raw input images. In digital pathology, most existing recurrent networks are trained with backpropagation and optimized with the cross entropy loss.

\subsection{Unsupervised Models}

A few works used unsupervised models to process digital pathology images. From the works we reviewed, the majority used auto encoders (or a variant) and generative adversarial networks in an unsupervised setting. 

Auto encoders (AEs) are feed forward neural networks that are formed of fully connected or convolutional layers with non linear functions used to compute each layer's activations. Most existing architectures follow the same design choices as the ones used in CNNs. Generally, AEs are composed of a contracting path that reduces the dimensionality of the input to a coarse latent representation and that is followed by an upsampling path that recovers the input. Thus, AEs are trained to reconstruct an input $\X$ on the output layer $\hat{\X}$ through one or multiple hidden layers $\h$. As AEs are used to reconstruct inputs, they are generally trained with the mean squared error which penalizes the reconstructed input $\hat{\X}$ for being dissimilar from the input $\X$. 
A sparse autoencoder (SAE) is simply an AE whose training criterion involves a sparsity penalty on the latent representation $h$, in addition to the reconstruction error. Stacked Sparse auto encoders (SSAE) are formed by stacking multiple auto encoder layers on top of each other and forming deeper networks.

In digital pathology, SSAE were used to obtain feature representations while leveraging datasets in an unsupervised setting (i.e., without requiring annotations). This feature learning strategy was shown successful on a variety of applications such as cancer identification~\cite{cruz2013deep, nayak2013classification} and nuclei detection~\cite{xu2016stacked}. 

Recently, another variant of AEs named variational auto encoders (VAE) was employed to generate grayscale histopathology images~\cite{tomczak2016improving}. VAEs are probabilistic models used as generative models which aim at learning a probability distribution over a given training dataset. To do so, VAEs are trained to maximize the variational lower bound on the log-likelihood of the data. These models are in essence different from general AEs but they do follow the same architectural design with a contracting path (the encoder) and upsampling path (the decoder). So far, there have been only rare applications of VAEs to digital pathology images.

Another form of generative models that was shown useful in digital pathology are the generative adversarial networks (GAN)~\cite{goodfellow2014generative}. The goal is to learn a generator distribution $P_G(\X)$ that matches the real data distribution $P_d(\X)$. GANs consist of optimizing a minimax game between a generator $G$ and discriminator 
network $D$. Both of these networks are generally designed as CNNs. $G$ generates samples from the generator distribution $P_G(\X)$ by transforming a noise variable $z \sim P_n(z)$ into a sample $G(z)$. $D$ aims at distinguishing samples from the true data distribution $P_d(\X)$ from generated samples $G(z)$. The optimization involves finding the parameters of both networks $G$ and $D$ using the following expression: 
\begin{equation}\label{eq:GAN}
    \min\limits_{G} \max\limits_D V(D,G) = \mathbb{E}_{x\sim P_d}\big[\log D(\X) \big] +
    \mathbb{E}_{z\sim P_n}\big[\log(1-D(G(z)))\big]
\end{equation}

Different variants of GANs have been used in digital pathology for tasks such as data augmentation, data harmonization, domain adaptation and staining normalization~\cite{senaras2018creating, janowczyk2017stain, bayramoglu2017towards, moeskops2017domain}. We discuss these works further in section~\ref{sec:stain}.

\section{Detection and Analysis of Histologic Primitives}

As mentioned previously, a critical prerequisite in the diagnosis of tissue sections is the analysis of histologic primitives such as cells, nuclei, lymphocytes, mitotic figures, etc. The detection of these important tissue components provides support for various quantitative analysis tasks including cell and nuclei morphology, such as size, shape, and texture which are biomarkers of abnormalities.  

There exist many challenges in the detection of histologic primitives in digital pathology. First, as mentioned in section \ref{sec:artefacts}, the appearance of tissue slides can exhibit background clutter introduced during image acquisition but there exist also significant variations on nuclei and cell sizes, shapes and intracellular intensity heterogeneity. Finally, cells and nuclei are often clustered into clumps and may overlap partially with one another. Challenges related to anatomical variations and artefacts make the task of detecting cell and nuclei particularly suited for deep learning models that can potentially learn robust feature representations. In this section we describe different deep learning models proposed for the detection of histologic primitives and specifically present supervised techniques with classification, regression and segmentation deep neural networks as well as unsupervised techniques that leverage deep auto encoders.

%% DETECTION FORMULATED AS CLASSIFICATION TASK

\subsection{Classification Models}

Many detection problems (i.e. nuclei, cell, mitosis detection) can be formulated as a pixel or patch-wise classification task and characterized by the appearance of objects to be detected. A common strategy for nuclei or cell detection is to train a CNN classifier as a pixel classifier~\cite{khoshdeli2017detection, liu2017novel, pan2015effective, xu2015efficient} where the network is trained in a supervised setting using patches centered around the object of interest. The trained CNN model is often a two-class classifier and can be applied in a sliding window fashion on WSIs to detect all histologic components of interest and output a probability map, where each pixel value indicates the probability of one pixel being at the center of an object. Therefore, the target objects can be located, in principle, by seeking local maximum in the generated probability map. This is generally followed by non-maxima suppression to improve the detection results. The advantage of using a deep learning model in this case is mainly in leveraging the learned feature representation that, if the training set allows, can be robust to variations such as rotation, staining appearance, etc. This approach is evidently more accurate than previous state-of-the-art hand-crafted features even when using relatively small architectures such as shown by a variety of works that adopted LeNet-5~\cite{lecun1998gradient} the 5-layer CNN model ~\cite{romo2016automated, wang2016subtype} or proposed 3-layer CNN classifiers to detect cells and nuclei~\cite{khoshdeli2017detection, khoshdeli2018feature}. 

One important limitation of such pixel-wise classification approach lies in their high computational costs. In practice, sliding window CNNs do not scale well to the million pixel WSIs. To overcome this problem, different strategies have been explored. Wang et al.~\cite{wang2016subtype} proposed to reduce redundant computations of neighbouring patches by introducing k-sparse convolution kernels, pooling and fully connected layers that are created by inserting zero rows and columns into the original kernels to enlarge the kernel size hence enabling the network to be trained with larger input tiles. The model is trained on $40\times40$ patches and tested on $551\times551$ tiles, hence speeding up the total computation time at test time. In an extension to this work~\cite{wang2016subtype}, the authors also showed how distributed GPU systems could speed up the total training time~\cite{wang2016deep}. Xu et al. showed the applicability of k-sparse kernels to cell detection on lung cancer images~\cite{xu2015efficient}. Giusti et al.~\cite{giusti2013fast} achieved a three order magnitude speedup compared to standard sliding window by performing convolutions with larger strides.

Another strategy proposed to improve CNN classifiers for detection tasks consists of utilizing domain knowledge about the object to be detected and leveraging the properties of the H\&E staining. In fact, when detecting nuclei or mitotic figures,  many works~\cite{wang2014mitosis, romo2017deep, romo2016automated, khoshdeli2018feature} observed that leveraging the hematoxylin channel which highlights nuclei in blue/dark purple would give a good estimate of the potential nuclei centers and hence can be leveraged to generate relevant candidate proposals to use along with the CNN for training. Often the hematoxylin image is filtered out by computing the laplacian of Gaussian (LoG) which allows to detect blob like structures then patches are extracted at those areas of detected nuclei to form candidate proposals~\cite{romo2016automated}. The LoG filter response can also be used to improve color decomposition when separating the hematoxylin and eosin channels~\cite{khoshdeli2018feature} which was shown to improve nuclei detection. Chen et al.~\cite{chen2016mitosis} proposed to use a FCN to first quickly retrieve mitosis candidates and output a score map indicating the probability of mitosis candidates. The FCN is followed by a CNN classifier in a cascaded fashion where the CNN receives the retrieved candidates for further classification into mitotic vs non mitotic nuclei. Both networks share convolutional layers and only the last fully connected layers differ resulting in two different outputs. The model is trained with a softmax classification loss and an L2 regularization term. Along the line of works proposing candidate proposals to use prior to detection; Akram et al.~\cite{akram2016cell} propose a two-branch CNN where one branch regresses on bounding box coordinates and the second predicts a classification score describing how likely the detected box is to include a cell. The generated candidate proposals are then used for training further training a cell classification CNN.

Pixel-wise classification can be easily extended to multi-class organ detection using a softmax loss when training the CNN model. Such multi-class CNN can still be trained with randomly sampled patches and applied on densely sampled patches at test time. A similar strategy was used by Cirecan et al.~\cite{cirecsan2013mitosis} to win the ICPR 2012 mitosis detection challenge where the task was to identify nuclei in mitotic phase within WSIs. While mitosis are normally rare and well-separated, they are very hard to differentiate from non-mitotic nuclei and the ratio of mitotic to non-mitotic nuclei is generally very low in a tissue slide. Hence, training datasets composed of randomly sampled patches generally suffer from high class imbalance. To overcome this problem, different techniques have been proposed. First, the most straightforward approach consists of sampling balanced datasets and using data augmentation strategies to augment the dataset for positive cases by replication as well as additional forms of augmentation based on affine transforms~\cite{malon2013classification}. Another approach is to leverage non-expert annotated datasets and learn from crowds. Albarquouni et al.~\cite{albarqouni2016aggnet} proposed AggNet, a CNN model for mitosis detection that can directly handle data aggregation in the learning process such that image annotation from non-experts can be leveraged during training. Specifically, AggNet consists of multi-scale CNNs that are first trained with expert annotations and tested on data. The output after this first round of training is sent to crowdsourcers for relabeling the mistaken annotated images. Then, the networks are refined using the collected annotations and used to generate new ground truth labels. 

Finally, another category of works leverages unsupervised learning. Xu et al.~\cite{xu2016stacked} proposed a SSAE for nuclei detection in breast cancer images. Their model is first trained in an unsupervised setting with a reconstruction loss and a sparsity constraint. Then, a softmax layer is added as penultimate layer and the model is fine-tuned to classify nuclei patches from non-nuclei patches. At inference, a sliding window method is employed to detect all nuclei within the new tissue slides. Song et al.~\cite{song2017hybrid}  proposed a hybrid deep auto-encoder to extract high level features from input patches and created probability maps that capture the different shape of nuclei using a Gaussian filtering of image patches centered on cells. The auto encoder is trained to reconstruct the input RGB image as well as the corresponding generated probability map. The predicted detected center points of nuclei are obtained by applying the deep auto encoder to unseen patches and using a second model, here a CNN or SSAE, to perform cell classification on the resulting predicted centers. In an extension~\cite{song2017simultaneous} to this first work, the classification CNN and the SSAE were coupled to create a model that first detects then classifies cells in WSIs. The coupling is done by adding two branches to the auto encoder: one classifies the detected cells and the other predicts probability maps.

%% ===== DETECTION FORMULATED AS REGRESSION TASK

\subsection{Regression Models}

Regression models for cell and nuclei detection were proposed to efficiently locate the centroid of objects. In fact, while CNN classification models produce relatively good results on detection tasks, they do not consider the topological domain on which the output detection resides.
In regression analysis, given an input and output pair, the task is to estimate a function that represents the relationship between both variables where the output does not only depend on the input but also on a topological domain (e.g. spatial domain, time).

Most works that address the detection of histology primitives with a regression model define a training output as a proximity score map, which indicates the proximity with respect to the centroid of the object of interest. Generally, the proximity score map is defined as a function resulting in high peak values in the vicinity of the center of the object. For instance, Chen et al.~\cite{chen2016automated} defined the proximity score map as:
\begin{equation}
    s(l) = \left\{\begin{matrix}
            0 & x \not\in \mathcal{M} \\ 
            e^{- \frac{||l-c||^2}{2\sigma^2}}& x\in\mathcal{M}
\end{matrix}\right. 
\end{equation}

\noindent where $l$ is the position of a pixel in the input patch with respect to the closest centroid $c$ of the object of interest (i.e. mitotic nuclei), $\mathcal{M}$ corresponds to a ground truth segmented mask composed of annotated objects and $\sigma$ is a hyperparameter controlling the variance.  

The model is trained with a per-pixel regression loss that minimizes the L2 norm between the predicted proximity score map and $s(l)$. In the deep regression CNN model, an FCN architecture is proposed with an upsampling layer that is used to predict a proximity score map of similar size as the ground truth map $s(l)$. Xie et al.~\cite{xie2015beyond} proposed a similar regression CNN for detecting cells in breast, cervex and neuroendocrine tumor tissues but used a weighted L2 loss formulated such that higher penalties are applied for errors on predicted pixels that are closer to the centroid of the cells.

Sirinukunwattana et al.~\cite{sirinukunwattana2015spatially, sirinukunwattana2016locality} introduced a new CNN layer designed for spatially constrained regression to detect epithelial tumor nuclei in breast and colon cancer datasets. The CNN predicts probability values that are topologically constrained such that high values are concentrated in the vicinity of the center of nuclei. In the spatially constrained CNN architecture, the two penultimate layers are defined to impose these spatial constraints. Specifically, the second to last layer outputs a parameter vector $\theta(x)$ while the final layer outputs a probability map $\hat{y}$ with highest values at the pixels that are strong candidates for being the centers of tumor nuclei such that:
\begin{equation} \label{eq:SCNN}
    \hat{y}_j \propto	 \left\{\begin{matrix}
                        \bigg(\frac{1}{1+ \frac{1}{2}||z_j-\hat{z}^0_m||^2}\bigg) h_m & \text{if} \quad \forall m \neq m' \quad ||z_j-\hat{z}^{0}_{m}||^2 \leq ||z_j - \hat{z}^{0}_{m'} ||^2 \\ 
                        0  & \text{otherwise}
                \end{matrix}\right. 
\end{equation}

\noindent where $z^0_m$ represents the coordinates of the center of the $m^{th}$ annotated nucleus within a given patch, $\hat{z}^{0}_{m}$ is the $m^{th}$ estimated nuclei center and $h_m \in [0,1]$ is the height of the output probability mask and is a trainable parameter defined as the output of the layer before last.
The model is trained with a weighted cross entropy loss that penalizes wrong predictions compared to a pre-computed ground truth regression mask. Finally, to detect the center of nuclei on large test image, the sliding window strategy with overlapping windows is used and the predicted probability of being the center of a nucleus is generated for each of the extracted patches using eq.(\ref{eq:SCNN}). These results are then aggregated to form a probability map where for each pixel location the probability values from all overlapping patches containing the given pixel are averaged. The
final detection is obtained from the local maxima found in the probability map. This spatially constrained regression CNN was improved in the work of Kashif et al.~\cite{kashif2016handcrafted} where the authors utilized a set of handcrafted features (i.e. color features after color decomposition to isolate hematoxylin channel and texture features based on the scattering transform~\cite{mallat2012group}) as additional inputs to the network and empirically showed sharper prediction results with lower false negative nuclei detection.  

To constrain the regression model, Xie et al.~\cite{xie2015deep} proposed a CNN-based approach for nuclei localization that mainly consists of two steps. First, patches extracted from a WSI are assigned a set of voting offset vectors that correspond to a preset number of voting positions, and a corresponding voting confidence score used to weight each vote. Then, the weighted votes are collected for all extracted patches and a final voting density map is computed for the input WSI. The final nuclei positions are identified by the local maxima of the density map. 
%%add the equations - maybe

Generally, detecting cells and nuclei is a first step towards counting or extracting quantitative measures such as nuclei size or total count of cellular structures in a WSI. Another direction for cell/nuclei analysis is to directly predict the desired measurement. Xie et al.~\cite{xie2018microscopy} proposed to bypass cell detection and directly estimate the total count of cells in histopathology images using a density estimation approach. Specifically, the cell counting problem is cast as a supervised learning problem that tries to learn a mapping from an input image patch to a density map which is a function over pixels in the image that is used to get an estimate of the number of cells in the image after integration. To learn this mapping, the authors proposed a fully convolutional regression network that is trained to regress ground truth density maps from the corresponding input image patches. The model is trained using the mean square error between the predicted heat map and the target density map. At inference, given an input image patch, the model predicts a density map. Two FCN architectures inspired by the VGG-Net architecture adapted with different receptive fields were tested. The authors observed that using larger receptive fields worked best when cells were organized as clumps (where multiple cells overlap) and hence covered larger image patches.

%% ========================================================================
\subsection{Segmentation Models}

Another approach for automating the analysis of histologic primitives is through analyzing the shape and morphology of cells, nuclei, glands or other tissue components. To do so, detection only is not sufficient and a pixel-level delineation of these components is better suited. However, segmentation in digital pathology is often challenging given the high resolution nature of WSI and deep learning models tend to be computationally inefficient on such large images. Most existing deep segmentation models are supervised models that leverage fully convolutional architectures trained with a per-pixel cross entropy loss. In fact, during the 2015 MICCAI Gland Segmentation Challenge~\cite{sirinukunwattana2017gland}, the top-5 teams used FCN architectures with different pre and post processing strategies. The challenge winning model~\cite{chen2017dcan} was a FCN model with auxiliary layers trained to simultaneously predict a gland boundary mask and a gland mask. Using the auxiliary loss functions as a form of regularization helped training a segmentation model that would converges to more plausible segmentation results in which gland boundaries are not clustered. Such additional domain specific knowledge is often critical when training deep models in digital pathology, especially given the scarcity of the dataset available. 
We discuss the most representative variants of FCNs in digital pathology below. 

Song et al.~\cite{song2015accurate, song2017accurate} proposed a multi-scale CNN architecture to segment cervical cytoplasm and nuclei. The WSI is partitioned into patches extracted at multiple scales and a CNN is trained for each scale to classify the central pixel of each patch as nuclei, cytoplasm or background. Feature maps from different scales are then concatenated and a final segmentation mask is obtained using a graph partitioning technique on the extracted features. At inference, the trained multi-scale CNN is applied in a sliding window fashion. Bel et al.~\cite{de2018automatic} also showed the advantage of using a multi-scale CNN model trained with patches extracted at two magnification levels (i.e. 20x and 40x) over a FCN for segmenting nine classes of renal structures.  Janowczyk et al.~\cite{janowczyk2018resolution} proposed a resolution adaptive deep hierarchical learning framework which uses multiple deep learning networks, with the same architecture, to significantly reduce computation time for fully segmenting high-magnification digital pathology images. 
Raza et al.~\cite{raza2017mimo} also used multiple resolution input patches to train a FCN for segmenting cells in fluorescence images but in contrast to other works, their proposed network uses intermediate layers in the network architecture to enforce better localization and context. The proposed network consists of multiple convolution followed by tanh activation, pooling and upsampling layers. Features from different layers are concatenated to enforce information flow from lower layers of the network to deeper layers. Auxiliary losses are added at different intermediate layers to penalize segmentation errors at cell borders. 

Another challenge in segmenting nuclei and cells in WSI is the potential overlap between neighbouring structures which can alter the cell/nuclei shape and boundaries appearance.  In order to preserve shapes many works rely on imposing higher order penalties when training segmentation models. While FCNs are generally trained with per-pixel cross entropy loss functions that do not necessarily encode specific shape priors, their output probability maps have been shown useful as unary terms in energy based segmentation models. 
Xing et al.~\cite{xing2016automatic} used such strategy to segment tumor nuclei from overlapping clumps and showed that the probability maps obtained from a trained FCN model formed a robust initialization step for a level set model.

As for detection, segmentation can also be bypassed to directly predict the desired quantification of histologic primitives. For instance, Veta et al.~\cite{veta2016cutting} proposed to avoid the nuclei segmentation step and directly predict the nuclear area which is predictive of outcome for breast cancer patients. The authors train a CNN model that is applied locally at each given nucleus location to measure the area of individual nuclei.

%%====================================================================================================
\section{Large tissue analysis and Prediction Models}

Often, the purpose of many tasks in digital pathology, such as counting mitoses, quantifying tumour infiltrating immune cells, analyzing the shape of glandular structures of specific tissue entities aim to ultimately predict patient outcome. Therefore, an interesting question is whether these intermediate proxies for outcome could be bypassed and the novel deep learning techniques could be used to directly learn diagnostically relevant features in microscopy images of the tumour, without prior identification of the known tissue entities, e.g., mitoses, nuclear shape, infiltrating immune cells. Different works attempted to train deep learning models using patient outcome as the endpoint to see if such model could reveal known prognostic morphologies, but also has the potential to identify previously unknown prognostic features~\cite{bejnordi2017deep}. We review some of these works in the following section.

\subsection{Deep Classification and Segmentation Models for Cancer Diagnosis}

Given a WSI, automatic cancer diagnosis is generally formulated as assigning a class label to an entire tissue slide (i.e., WSI classification) or identifying abnormal areas of tissues (i.e., WSI segmentation). The majority of existing deep models proposed for predicting the presence of cancer from WSIs use CNNs and FCNs. 

There are three main challenges in designing CNN models for cancer prediction from WSIs. All of these challenges arise from the high dimensionality of tissue slides and the limited available annotations for training and validating supervised models. As mentioned in section~\ref{sec:data}, patch-based representations are unavoidable but come with different shortcomings. In this section, we discuss the different strategies proposed for i) training supervised patch-based prediction models using fine-level annotations (i.e., delineations of abnormal tissue areas in a WSI), ii) training weakly supervised patch-based prediction models using only slide-level annotations (i.e., benign vs malignant WSI), iii) encoding the topology of tissues by leveraging the multi-magnification nature of WSIs. 

\subsection{Fully Supervised Models}

In fully supervised settings, cancer prediction models can be trained on finely annotated datasets in which one (or several) expert(s) highlighted abnormal areas (e.g., cancerous tissue) within a WSI. In fact, in most applications, only a small portion of the imaged tissue caries morphological and structural patterns indicative of abnormalities. Delineating such areas can be very complex and time consuming as it involves visually identifying abnormal patterns at different magnification levels. 

Different works proposed CNN classifiers trained on patches extracted from annotated WSIs to predict the presence of cancer or cancer subtypes. In most works, a fixed magnification level is selected (generally 20x or 40x) and patches are sampled from the set of annotated training WSIs. Sampling strategies in cases where pixel-level annotations are available, involve constructing training mini-batches by randomly selecting samples from points inside and outside the annotated area but within the tissue (i.e., discarding background). For each mini-batch the number of samples per class is generally determined with uniform probabilities~\cite{bejnordi2017diagnostic, bejnordi2017deep, qaiser2018her, cruz2017accurate} and the patch size is often large enough to include structural information (e.g., $224\times224$). This strategy can work poorly and result in very imbalanced mini-batches in cases where only a fraction of the tissue contains abnormality. This can be the case, for instance, for breast cancer metastasis detection in sentinel lymph nodes where often, only a minuscule fraction of the slide contains cancer and most of the slide is covered by lymphocytes. Also, certain normal regions which look more similar to cancer are typically underrepresented in the training data and patch-based CNN models are generally not capable of correctly identifying these areas as normal.  To address this problem, Litjens et al.~\cite{litjens2016deep} used a boosting approach to sample training mini-batch patches. This consists of using the initial prediction score maps obtained for the training dataset to sample new patches for both cancerous and non-cancerous classes while increasing the likelihood of sampling patches that were originally incorrectly classified by the model. This process results in additional training data which contains more difficult samples. Subsequently, the model is re-trained using the new sampled patches. 

Once the patch-level classification model is trained, it can be used in a sliding window fashion to classify all patches in an unseen WSI and obtain a prediction score map highlighting areas predicted as cancerous. To obtain a slide-level prediction for the entire WSI from the predicted score map, different aggregation techniques have been proposed~\cite{liu2017detecting}. Some works simply use the maximum value in the predicted probability score map as the slide-level prediction~\cite{xu2016deep, liu2017detecting}. One strategy that was shown successful on different datasets~\cite{golden2017deep, qaiser2018her, bejnordi2017diagnostic, graham2018classification, bejnordi2017deep} consists of leveraging domain knowledge to extract additional hand-crafted features from the predicted score map and train a subsequent classifier to predict a slide-level label. Wang et al.~\cite{wang2016deep} adopted this strategy to win the Cameleyon~\cite{bejnordi2017diagnostic} challenge on identifying metastatic breast cancer from lymph node WSIs. Specifically, 28 geometrical and morphological features were extracted from each predicted score map, including the percentage of tumor region over the whole tissue region, the area ratio between tumor region and the minimum surrounding convex region, the average prediction values, and the longest axis of the tumor region. A random forest classifier is then trained to discriminate WSIs with metastases from the ones without using the hand-crafted features.

%=========
\subsection{Weakly Supervised Models}

Weakly supervised models leverage patch-based representation to classify WSIs while only using slide-level annotations during training. Among the works we surveyed, there exist different strategies for training a prediction CNN model using slide-level labels, we present them below. 

One commonly used strategy consists of extrapolating the labels from the slide-level to all patches sampled from a given WSI, train a patch-level classifier then use max-voting or pooling to infer the slide-level class. This strategy can be used for patch-level segmentation with FCNs~\cite{qaiser2017tumor} as well as WSI classification~\cite{bejnordi2017diagnostic}. Depending on the complexity of the task and the disparity among patches, this strategy can produce relatively good results. More sophisticated approaches adopt the multiple instance learning framework~\cite{jia2017constrained, hou2016patch}. For instance, Hou et al.~\cite{hou2016patch} aggregate the prediction of a patch-based CNN model using an Expectation-Maximization (EM) based decision fusion model to automatically locate discriminative patches. In the authors' formulation, a WSI $\X^{(i)}$ is represented as a bag (using the MIL terminology) of instances (i.e., patches) $(\x^{(i)}_1, \x^{(i)}_2, \dots, \x^{(i)}_P)$ where $P$ is the total number of instances per bag. Ground truth labels are only available at the bag level.  Hidden binary variables $H^{(i)}$ are introduced to model whether an instance of a bag is discriminative or not. Each instance is associated with a hidden variable such that $h^{(i)}_p$ represents the hidden variable associated with patch $p$. Assuming all bags and instances are independent, the decision fusion model is defined as follows:

\begin{equation}
    P(\X, H) = \prod\limits_{i=1}^N \prod_{j=1}^P \bigg( P(\x_{j}^{(i)}| h^{(i)}_j)\bigg) P(H^{(i)}) ,
\end{equation}
\noindent where $N$ is the total number of training WSIs. The likelihood of the data $P(\X)$ is maximized using the EM algorithm and the instance that maximizes  $P(h^{(i)}_j | \X^{(i)})$ is defined as the discriminative instance for the positive bag. Discriminative instances are selected to continue training the CNN prediction model.

Nazeri et al.~\cite{nazeri2018two} proposed a two-Stage CNN for classifying breast cancer WSIs. A first CNN, called patch-wise network is trained on patches and outputs spatially smaller feature maps. A second CNN network (named image-wise network) which performs on top of the patch-wise network, receives stacks of feature maps as input and generates slide-level prediction scores. In this cascaded configuration, the second network learns relationships between neighbouring patches represented by their feature maps. The patch-level labels are unknown yet the patch-wise network is trained using the CE loss based on the label of the corresponding WSI. Once trained, the last fully connected layer of the patch-wise network are discarded and only the feature maps obtained from the last convolutional layer are used. The image-wise network is trained on the feature representations obtained from non-overlapping patches using the CE loss again.

Other works proposed to use domain knowledge to identify discriminative patches without requiring patch-level annotations. For instance, Ertosun et al.~\cite{ertosun2015automated} observed that the grading of gliomas is highly correlated with nuclei morphology, hence, tissue areas with abundant nuclei distribution should be more reflective of the slide-level glioma grade. Based on this observation, the authors trained a CNN model on patches extracted after automatic nuclei detection and showed improved performance against randomly sampled patches.

%===========
\subsection{Encoding Structural Information}

Capturing high-level contextual information with patch-based representations is generally difficult as it involves handling larger input image sizes, which does not work well when training deep models. In order to capture context, different works attempt to encode structured relations between neighbouring patches. 

One strategy to encode larger context is to employ larger input patches when training the deep prediction models. This strategy was explored by Bejnordi et al.~\cite{bejnordi2017context} witch stacked multiple CNN models trained sequentially with input patches of increasing size in order to learn fine-gained (cellular) information and global interdependence of tissue structures. The authors leveraged the FCN architecture to train the model with patches of increasing size. Specifically, a first CNN model is trained to classify patches of size $224\times224$. After convergence, this first model is freezed and converted into a FCN architecture by removing the last fully connected layers. A second network is then stacked on top of the FCN and is similarly trained to classify patches. The input to the second network are the feature maps output of the first FCN. In this cascaded architecture, the second model is trained with input patches with larger size.

Other strategies to encode larger context involve using LSTM units. Agarwalla et al.~\cite{agarwalla2017representation}	combined the different feature representations obtained from a patch-level CNN using a 2D LSTM network to encode neighbouring relationships between patches. A 2D-grid of features is generated by packing feature vectors for neighbouring patches in the WSI then four 2D LSTMs running diagonally are used to capture the context information by treating WSIs as a two-dimensional sequence of patches. Finally, tumour predictions across all the spatial dimensions are averaged together to get the final slide-level class label. Similarly, Kong et al.~\cite{kong2017cancer}	combined CNNs and 2D LSTMs to encode structure between neighboring patches for classifying metastatic tissue slides. The spatial dependencies between patches are explicitly modelled via an additional custom loss function which penalizes the CNN from predicting diverging probability scores for neighbouring patches in a 4-connected neighbourhood.

Finally, Wang et al.~\cite{wang2017adversarial}	attempt to encode structural information related to the contours of microinvasive cervix carcinoma regions in WSIs. Explicitly formulating domain knowledge related to membranes organization is difficult, especially when such information needs to be encoded within a deep learning model. Wang et al. ~\cite{wang2017adversarial} trained a segmentation FCN model  to segment basale membranes of cervix carcinoma tissues. In order to augment the FCN with additional information related to the organization and structure of the membranes' contours, the authors leverage adversarial training and include a GAN model in which the generator is replaced by the FCN segmentation model and the discriminator learns to identify ground truth membrane contours from predicted segmentation contours.

% =========================================================================
\subsection{Survival Prediction Models and Multimodal Applications}

Survival analysis is the task of predicting the time duration until an event occurs, which, in digital pathology, corresponds to the death of a patient. In survival datasets, each patient $(i)$ is labelled with a pair $(t^{(i)}, \delta^{(i)})$ corresponding to an observation time and a censored status. A censored patient (i.e., patient for which the event is not observed) is characterized by an indicator variable $\delta^{(i)} = 0$ while an uncensored patient (i.e., patient for which the event occurred during the study) is characterized by $\delta^{(i)}=1$. The observation time $t^{(i)}$ can be either a survival time $S^{(i)}$ or a censored time $C^{(i)}$ determined by the status indicator variable $\delta^{(i)}$. The most popular survival model is Cox proportional hazard model~\cite{cox1992regression} which is built on the assumption that a patient's survival risk is a linear combination of covariates (e.g., structured data such as patients' sex, smoking years, age).  However, linear models are not the best suited to model interactions in real-world datasets.  This motivated research that leveraged the non-linear deep learning models as survival predictors. We present some of the recent works below.

One way to leverage deep learning models in survival prediction tasks is to employ CNN features to discover new imaging biomarkers. For instance, Yao et al.~\cite{yao2016imaging} trained a CNN model to identify cancer from non-cancer cells in tissue patches and used the features from the last convolutional layer of the CNN to extract quantitative descriptors such as textures and geometric properties (e.g., cell area, perimeter, circularity).  The extracted features are used to discover imaging biomarkers that correlate with patient survival outcomes using the multivariate cox proportional Hazard model. The authors showed that imaging biomarkers from subtype cell information can better describe tumor morphology and provide more accurate prediction than other techniques relying od molecular profiles. Similarly, Bychkov et al.~\cite{bychkov2016deep} showed that CNN-derived features extracted from segmented images of epithelium, non-epithelium, and unsegmented tissue micro-array cores correlated well with five-year survival. Bauer et al.~\cite{bauer2016multi} showed that a CNN model could be used to classify nuclei in different tissue types (i.e., prostate cancer and renal cell cancer) and that the combination of tissue types during training could increase the overall survival analysis. Zhu et al.~\cite{zhu2016lung} showed that the integration of CNN-based features with genomic features into a Cox survival prediction model revealed complementary information on tumor characteristics between pathology images and genetic data.

Other methods employ CNNs to directly predict the survival risk factor. Zhu et al.~\cite{zhu2016deep} developed the first deep CNN for survival analysis (DeepConvSurv) with pathological images. The model is trained on patches extracted from tissue areas delineated by experts to regress the survival time for each patch.

\begin{equation}
    \L = - \sum\limits_{i\in U} \bigg(\beta^T a^{(i)} - \log \sum\limits_{j\in \omega^{(i)}} e^{\beta^T a^{(j)}}\bigg),
\end{equation}
\noindent where $\beta^T a^{(i)}$ is the risk associate with the input image with $\beta$ corresponding to the weights of the final fully connected layer and $a^{(i)}$ are its input activations. $U$ and $\omega^{(i)}$ represent the set of right-censored patients (i.e., cases for which the event did not occur at the time of the study) and the set of patients for which the event occurs after $t^{(i)}$. 

The authors extended this model to predict survival outcomes for an entire WSI (i.e., a collection of patches)~\cite{zhu2017wsisa}. In this extension, multiple patches are extracted from a given patient's WSI and clustered based on their phenotype with a k-means clustering approach. Then, clusters are selected based on  a their patch-level survival prediction performance using the Cox regression model for predicting survival. The selected clusters are then used to train the DeepConvSurv model described above. All cluster-level survival predictions are then aggregated to obtain a final survival prediction score for an entire WSI. 

In practice, patient diagnostic is generally based upon the integration of different sources of information (e.g., omics, radiology, patient history). Consequently, it may be beneficial to integrate these different sources of information into the automatic prediction systems as well. Recent works have attempted to design survival prediction models that integrate both genomic biomarkers and digital pathology images. Yao et al.~\cite{yao2017deep} used a survival CNN model trained on such multi-modal inputs to predict time-to-event outcomes. Their framework showed superior results to the clinicians' decision rules for predicting the overall survival of patients diagnosed with glioma. Given the current trend in personalized medicine, such multi-modal approaches are becoming more and more critical.

%%====================================================================================================
\section{Non-clinical Tasks}

In this section we present categories of work that do not directly attempt to automate clinical tasks but rather focus on facilitating computer-aided diagnosis by improving the quality of input images and datasets to facilitate training or introduce ways to design more interpretable deep learning models. 

\subsection{Stain Normalization, Computational Staining and Augmentation} \label{sec:stain}

Stain normalization has always been a critical step in the training of machine learning models. In fact, in contrast to pathologists, most existing machine and deep learning models are sensitive to variations in the staining appearance of tissue slides. 

Deep learning models were applied in various ways to normalize stains across datasets. Janowczyk et al.~\cite{janowczyk2017stain} proposed to use features learned with a SAE to normalize stains in different images to a template image. This process was applied by clustering the pixels in a sparse auto-encoded feature space so that respective tissue partitions could be aligned using their respective color distributions. By using individual tissue partitions, this approach is able to more sensitively modify the color space as compared to a global method where all pixels are considered concurrently. 
GANs have also been employed to normalize stains. Zanjani et al.~\cite{zanjani2018stain} trained a conditional GAN to generate a colorized H\&E image in its CIEL*a*b* space using as input the image lightness channel and a set of structured latent variables drawn randomly from a prior distribution. Finally, Cho et al.~\cite{cho2017neural}	defined the problem of stain normalization in the context of domain adaptation and combined it with a prediction task within a GAN framework. The GAN loss described in eq.(\ref{eq:GAN}) is combined with two additional regularization terms that aim at preventing the degradation of a task-specific network on synthetic images and enforce similar features between synthetic and original images in order to facilitate the task of the generative model. 

%\cite{schuler2017context}	Context-based Normalization of Histological Stains using Deep Convolutional Features:  image-to-image translation. To approach the problem of translating H&E-stained WSIs to their IF counterparts, we have applied the cGAN-driven algorithm pix2pix,10 which benefits from its bipartite formulation. Like other methods proposed for image-to-image translation, cGANs learn a functional mapping from an input image x to translated image y, i.e. G : x ↦ y, but, unique to a cGAN framework, it is the task of a generator G to generate the image y conditioned on x that fools an adversarial discriminator D, which is in turn trained to tell the difference between ground truth and generated images. In order to balance sensitivity and specificity in this context, we hypothesize that a generative model can be receptively tuned to encode sparse staining by being maximally penalized when it makes false classifications on low-prevalence ground truth tiles during training. Thus, we propose a prevalence-based adaptive regularization parameter λ′ that may be more suitable for the translation of signals from H&E to IF:

Conditional GANs were also used in the context of digital staining where the task is to generate images artificially stained with different staining agents. Burlingame et al.~\cite{burlingame2018shift} used a GAN to convert H\&E images to immunofluorescence. 	Bayramoglu et al.~\cite{bayramoglu2017towards} employed a similar conditional GAN to virtually stain unstained hyperspectral specimens. 

Finally, GANs have also been employed as data augmentation tools to generate synthetic images. Senaras et al.~\cite{senaras2018creating} and Moeskops et al.~\cite{moeskops2017domain} used a GAN to generate synthetic images and showed that using the synthetized images to augment training datasets increased the performance of prediction and detection models when compared to other standard augmentation strategies.

\subsection{Interpretability}

One recurring critique made against deep learning and machine learning models is their lack of interpretability. Generally, interpretable models in digital pathology are ones for which the output is jutifiable. Deep learning models do not a priori fit in this category and are commonly seen as black-box computational tools. 
Introducing novel ways to gain insight into the decision making process of deep learning models has become an active area of research in digital pathology.

Cruz-Roa et al.~\cite{cruz2013deep} proposed to interpret the predictions of a CNN used to identify cancer in WSIs. Their strategy for interpretability consists of including an interpretable layer that highlights the visual patterns contributing to discriminate between cancerous and normal tissues patterns, working akin to a digital staining which spotlights image regions important for diagnostic decisions. This is achieved by multiplying the feature maps of the trained CNN model by the final layer's softmax classifier weights. All weighted feature maps are combined into an integrated feature map and a sigmoid function is applied at each pixel of the resulting map. The map is used to highlight areas of the image that were scored as highly discriminative of cancer. While this approach gives a relatively good insight into the trained models' predictions, it does not really capture the intermediate layers' contributions to the final outcome. 

Korbar et al.~\cite{korbar2017looking} proposed a visualization approach to identify discriminative features for colorectal polyps in WSI. In their work, the FCN nature of ResNets is leveraged to directly correlate features with salient areas at each layer. Specifically, the authors used a gradient-based approach to identify discriminative features learned at each intermediate layer of the trained ResNet model. Gradients reflect the change of each intermediate layer function with respect to the input. Hence, visualizing the resulting gradient maps can provide relevant insight on the features distribution per class. However, gradients have generally high variance at intermediate layers and they can be influenced by all different output classes. The authors used a variant of gradient based visualization technique name the class activation map which looks at the change in the penultimate layer's activations after a single backpropagation pass. A class activation map for a particular class indicates the discriminative image regions used by the CNN to identify that class. The utility of this approach in identifying areas of the input image that justify the most the models' prediction was tested by comparing the obtained class activation maps with expert annotations. The method showed promising results and relatively good estimates of decisive regions and features for different types of polyps were obtained.

These techniques are a first step toward gaining more insight on the training and output predictions obtained from deep learning models. At a clinical level, interpretability offers a potential avenue for introducing deep learning models into clinical workflows, hence, we can expect future works to continue pursuing this area of research.

%\subsection{Image Retrieval Systems}

%%====================================================================================================
\section{Validation Strategies}

A wide variety of metrics have been used to validate the different deep learning models. In the works we reviewed, accuracy, F1 score, precision, recall and area under ROC curve (AUC) were used in majority to evaluate detection and prediction models. Depending on the application, the performance can be reported at the patch or slide-level. Dice scores and FROC are generally used for evaluating segmentation models and measuring the overlap between the predicted segmentation mask and the ground truth annotation.  Veta et al.~\cite{veta2016cutting} used the Bland-Altman method to evaluate the agreement between two sets of measurements (i.e., predicted vs measured nuclei area).  For generative models used in stain normalization tasks, there was no common metric used among the works we surveyed. In fact, some works chose to evaluate the color constancy of the stain normalized images and used the normalized median intensity, others focused on perception-based metrics and reported the accuracy of experts attempting to differentiate synthetic from real images after stain normalization. 

To evaluate the performance of prediction models, a few works attempted to compare the performance of the proposed deep learning models with experts on similar test sets. Different metrics have been used to compare humans to machines. For instance, agreement between experts and automatic systems is ideally quantified using the Kappa score.  Another approach involves comparing the AUC of an automatic system to the average sensitivity and specificity of different experts. While this is not a completely fair comparison, it gives a relatively good estimate of how far are current systems from becoming part of clinician's workflow.

With the increase of available public datasets and the emergence of many challenges and competitions, the use of cross-validation techniques to evaluate machine learning systems has reduced. In fact, most released datasets as part of competitions are split into training, validation and test sets and these splits are generally kept fixed to facilitate comparison between the competing methods. This strategy, however, does not allow models to be tested on different training sets. 

There exist different types of annotations used in the datasets surveyed in this report. Most datasets were annotated by one (or more) pathologist at the slide or patch level and only a few public datasets provide annotations confirmed by immunohistochemistry, genetics or patient's outcome. Annotations obtained from experts can be imperfect, especially in tasks such as segmentation of cancer areas where human subjectivity is unavoidable. Also, appropriate use and interpretation of annotations collected from multiple experts remains an open challenge. The gold standard in digital pathology is generally seen as the survival outcome of the patient or their molecular profiling. However, a model that predicts cancer from patient's survival unavoidably makes strong assumptions regarding the causal relationship between the patient's observational data (i.e., a digital pathology slide) and their outcome. Verifying the statistical prevalence of such relationship is impossible.

With the increasing number of large dataset collections made publicly available, we can expect evaluation metrics and validation strategies to become more and more standardized.

%%====================================================================================================
\chapter{Summary and Discussion}

Since the adoption of the digital slide for clinical diagnosis there has been a gradual evolution over the years aimed at reducing manual intervention and automating pathologists' workflow. In the initial phase of digital pathology, traditional computer vision methodologies were used for tissue detection, segmentation, morphometry, and a plethora of other tissue analysis tasks. The main challenges for designing accurate systems are the variability in staining of tissue slides, slide preparation and artefacts resulting from the digitization process.

One particularity of the WSI is its high dimensionality which allowed for early applications of deep learning models with patch-based representations as it provided large enough training sets obtained by sampling patches from the large tissue slide. The studies presented in this survey report the state-of-the-art performance on most WSI analysis tasks (i.e., analysis of histology primitives and outcome prediction) and are all based on (or leverage) patch-based deep learning models. These models have enabled accurate prediction models as well as techniques for identifying and extracting discriminative information from complex tissue images.

While computational imaging with deep learning models can clearly play a role in better quantitative characterization of disease and precision medicine, there still remain a number of substantial technical and computational challenges that need to be overcome before computer assisted image analysis of digital pathology can become part of the routine clinical diagnosis. Although some of the existing deep learning models were designed to overcome challenges related to the WSI size (e.g., patch-based strategies, efficient sampling techniques, faster operations via improved hardware or sparser models), processing the multi-magnification tissue slide without loss of context or structural information remains unsolved. In fact, most of the models employed in the current works are not designed for large input sizes. There is also a need for more standardized and clinically relevant validation protocols as currently none of the existing works were tested on real clinical cohorts. On a similar note, designing automatic cancer diagnosis systems is still an open problem that arises from the lack of available annotated datasets as there is generally no clear clinical consensus on subtyping cancers. Ultimately, the clinical applicability of any deep learning based prediction system will unavoidably result from a collaborative effort between pathologists and computational scientists in order to clearly identify relevant clinical problems and accurately interpret the available annotations.

Aside from improving the performance and interpretability of existing systems, there are many novel directions that could be explored. With the availability of larger datasets such as the TCIA~\cite{TCIA} and TCGA~\cite{TCGA} data sharing portals, models that can leverage a fusion of modalities (e.g., genomics, histopathology, radiology) for better predictive modelling could be an application area for deep learning. There are also exciting opportunities in leveraging deep models to improve the digital slide acquisition procedure with learning based approaches for signal reconstruction but also flagging systems that can identify altered digital slides  (due to staining, fixation or any other artefact caused by the image acquisition) and alleviate the need for manual interventions during the digitization process. On the technical side, the emergence of new datasets such as PCam~\cite{pcam} which is the first dataset for digital pathology that includes more than 300,000 images of size $32\times32$, brings new opportunities for developing and evaluating customized deep network architectures that are inspired by the challenges proper to tissue images. For instance, one active area of research in this direction is the design of rotation invariant CNN models.

To conclude, by all indications, the digitization of tissue glass slides and the large adoption of deep learning models as computation models is showing significant potential for a transformation of the field of digital pathology from qualitative to quantitative.

%   BACK MATTER  %%%%%%%%%%%%%%%%%%%%%%%%%%%%%%%%%%%%%%%%%%%%%%%%%%%%%%%%%%%%%%
%
%   References and appendices. Appendices come after the bibliography and
%   should be in the order that they are referred to in the text.
%
%   If you include figures, etc. in an appendix, be sure to use
%
%       \caption[]{...}
%
%   to make sure they are not listed in the List of Figures.
%

\afterpage{%
{\small
    %\clearpage% Flush earlier floats (otherwise order might not be correct)
    %\thispagestyle{empty}% empty page style (?)
    %\begin{landscape}% Landscape page
        \begin{table}[t] %% 
            \centering
            \caption{Public Datasets for Image Analysis Tasks in Digital Pathology. }
            \begin{adjustbox}{width=0.7\textwidth,totalheight=0.8\textheight,center=\textwidth,keepaspectratio}
            \begin{tabular}{ l l l}
            \toprule
            Dataset & Task & \#Images  \\
            \midrule 
            
            Arganda et al.~\cite{arganda2015crowdsourcing} & Neuron boundary segmentation & 60 \\
            ISBI2012-EM~\cite{em} & Neuron segmentation & 30 \\ 
            GlaS~\cite{sirinukunwattana2017gland} & Colon gland segmentation & 160 \\  
            AMIDA2013~\cite{veta2015assessment}& Mitosis detection &   \\
            ICPR 2012~\cite{cirecsan2013mitosis} & Mitosis detection & 50 \\
            ICPR 2014~\cite{cirecsan2013mitosis} & Mitosis detection & 50 \\
            TUPAC~\cite{tupac} & Tumor detection & 300  \\
            Camelyon16~\cite{bejnordi2017diagnostic} &Metastasis detection  & 299 \\
            Camelyon17~\cite{bejnordi2017diagnostic} &Metastasis detection  &  400\\
            TMA Thyroid~\cite{TMAThyroid} &Outcome prediction  &  \\
            BACH~\cite{BACH}& Tissue subtypes classification& 400   \\
            TCIA~\cite{TCIA} &Multiple & --  \\
            TCGA~\cite{TCGA} &Multiple & --  \\
            Her2~\cite{qaiser2018her}& Her-2 Scoring &  172   \\
            Cellavision~\cite{cellavis}& Cell segmentation &100   \\
            Enjoypath~\cite{enjoypath}& Multiple& 318     \\
            PCam~\cite{pcam} &Metastasis detection &327680   \\
            \bottomrule
            \end{tabular}
            \end{adjustbox}
            
            \label{table:datasets}
            \end{table}
        %\captionof{table}{Table caption}% Add 'table' caption
    %\end{landscape}
    }
    \clearpage% Flush page
}

%{\footnotesize

\begin{landscape}

\tiny
\setlength\LTleft{0pt}
\setlength\LTright{0pt}
{\setlength{\tabcolsep}{0.7pt}
\begin{longtable}{@{\extracolsep{\fill}}lp{1.2cm}p{0.8cm}p{0.8cm}p{2cm}p{0.8cm}p{0.9cm}p{2cm}p{1.2cm}p{1.5cm}p{1.5cm}
p{1cm}p{1cm}p{1cm}p{1cm}p{1.8cm}}

%\begin{longtable}{p{1.3cm} p{1cm} p{1cm} p{1cm} p{2cm} p{1cm} p{1cm} p{2cm} p{2cm} p{2cm} p{2cm} p{1cm} p{1cm} p{1cm} p{1cm} p{1cm} p{1cm}}

\caption{Deep Learning Models for Digital Pathology Applications. The nomenclarture used in this table is as follows. PP:Pre-processing, Augm.:Augmentation, Annot.: annotations, CONV:convolution layer, MP: max pooling, ReLU: rectified linear units, FC: fully connected layer, BN: batch normalization, UP: upsampling, ELU: exponential linear unit, LRN: local response normalization, lReLU: leaky ReLU, ADV: adversarial loss, ConfMat: Confusion matrix, TPV: true predictive value, PPV: positive predictive value. Values separated by "/" correspond to different datasets used in the corresponding study.  }
\label{table:sota} \\
\toprule
    
Method & Site & Stain & Mag. & Task & \#WSI & \#Annot.  & Public Data & Input Size & PP & Augm. & Model & \#Layers & Ops & Cost  &  Performance   \\

\midrule
\endfirsthead
%\multicolumn{16}{l}{\footnotesize\itshape\tablename~\thetable} \\
\toprule
Method & Site & Stain & Mag. & Task & \#WSI & \#Patches  & Public Data & Input Size & PP & Augm. & Model & \#Layers & Ops & Cost  &  Performance   \\
\midrule
\endhead

\midrule
%\multicolumn{16}{r}{\footnotesize\itshape\tablename~\thetable} \\
\endfoot
% piede finale
\bottomrule
%\multicolumn{16}{r}{\footnotesize\itshape\tablename~\thetable} \\
\endlastfoot

%\multicolumn{16}{|c|}{HISTOLOGY PRIMITIVES}\\
\cite{cirecsan2013mitosis}&Breast&H\&E&40x&Mitosis detection&50&300&ICPR2012&$101\times101$&--&Rotations, Mirroring, Flip&CNN&13&CONV, ReLU, MP, FC&CE&P=0.88, R=0.70, F1=0.78\\

\cite{malon2013classification}&Breast&H\&E&40x&Mitosis detection&50&300&ICPR2012&$72\times72$&--&Rotations &LeNet-5&7&CONV, Tanh, MP, FC&CE&F1=0.66\\

\cite{wang2014mitosis}&Breast&H\&E&400x&Mitosis detection&50&300&ICPR2012&$80\times80$&--&Rotations, Mirroring&CNN&3&CONV, ReLU, MP, FC&CE&F1=0.73\\

\cite{chen2016mitosis}&Breast&H\&E&40x&Mitosis detection&50&300&ICPR2014&$94\times94$&--&--&FCN&16&&CE&F1=0.79\\

\cite{albarqouni2016aggnet}&Breast&H\&E&40x&Mitosis detection&23&--&AMIDA2013&$33\times33$&--& Rotation, Mirroring,Crowd&CNN&5&CONV, ReLU, MP, FC&CE&AUC=0.86, F1=0.61\\

\cite{romo2017deep}&Breast&H\&E&40x&Mitosis detection&174&&AMIDA&$64\times64$&--&--&CNN&5&CONV, ReLU, MP, FC, BN&CE&F1=0.56\\

\cite{romo2016automated}&Breast&H\&E&20x&Tubule detection&174&7513&--&$64\times64$&--&--&CNN&5&CONV, ReLU, MP, FC&CE&F1=0.59, P=0.72, R=0.56 \\

\cite{khoshdeli2017detection}&Brain Breast&H\&E&40x&Nuclei detection&29&13766&--&$51\times51$&SN&--&CNN&3&CONV, ReLU, MP, FC&CE&P=0.69, R=0.74, F1=0.72\\

\cite{xu2016stacked}&Breast&H\&E&--&Nuclei detection&535&3500&--&$34\times34$&--&--&SSAE&2&FC&CE&AP=78.83\\

\cite{khoshdeli2018feature}&Brain Breast&H\&E&--&Nuclei detection&29&13766&--&$51\times51$&--&--&VGG&3&CONV, ReLU, MP, FC, DO&CE&F1=0.84, R=0.81, P=0.88 \\

\cite{akram2016cell}&Blood Brain Bone Marrow&Fluo&40x&Cell detection&184&--&Fluo-HeLa&$53\times53$&--&--&FCN&7&CONV, ReLU, MP, FC&CE&IOU=0.72\\

\cite{wang2016subtype}&Lung&H\&E&--&Cell detection&300&--&--&$40\times40$&--&--&LeNet-5&7&Sparse-CONV, ReLU, MP, FC&CE&F1=0.82\\

\cite{liu2017novel}&Lung Neurons&H\&E&40x&Cell detection&16/24&150&--&$31\times31$&--&Rotations&CNN&5&CONV, ReLU, MP, FC&CE&F1=0.90/0.92\\

\cite{pan2015effective}&Lung&H\&E&--&Cell detection&215&83245&NLST&$20\times20$&--&--&CNN&7&CONV, ReLU, MP, FC&CE&F1=0.79\\

\cite{xu2015efficient}&Lung&H\&E&40x&Cell detection&215&83245&NLST&$20\times20$&--&--&LeNet-5&5&CONV, ReLU, MP, FC&CE&F1=0.79\\

\cite{xu2016detecting}&Lung&H\&E&40x&Cell detection&215&83245&--&$20\times20$&--&--&AE&2&FC&CE&F1=0.83\\

\cite{song2017hybrid}&Bone Marrow&H\&E&--&Cell detection&52&5248&--&$29\times29$&CD&--&AE& 8 &FC&CE&P=0.92, R=0.97, F1=0.94\\

\cite{song2017simultaneous}&Bone Marrow&H\&E&--&Cell detection&52&5248&--&$29\times29$&CD&--&AE&2&FC&CE&P=0.92, R=0.97, F1=0.95\\

\cite{gao2017hep}&Hep2&IF&--&Cell classification&83&13596&ICPR2014&$78\times78$&--&Rotation&AlexNet&7&CONV, ReLU, MP, FC, DO&CE&mACC=0.88\\

\cite{han2016hep}&Hep2&IF&--&Cell classif.&83&10000&ICPR2014&$78\times78$&--&Affine&CNN&5&CONV, ReLU, MP, FC&CE&ACC=0.96\\

\cite{zhao2017automatic}&Blood&H\&E&40x&LKC classif.&18&1080&ISBI2012-EM&$ 11\times 11$&--&None&CNN&4&CONV, MP, FC&CE&ACC=0.93\\

\cite{mishra2016structure}&MTC&EM&20Kx&MTC classif.&403&403&ICPR2012&$400\times400$&--&Rotations&LeNet-5&7&CONV, Tanh, SPM, FC&CE&CS=100\\

\cite{malon2013classification}&Breast &H\&E&40x&Mitotic grade&5&226&AMIDA2013&$72\times72$&--&None&CNN&10&CONV, ReLU, MP, FC&CE&F1=0.66\\

\cite{giusti2013fast}&Neuron&EM&--&Neuron segm.&30&--&ISBI2012-EM&$95\times95$&--&--&CNN&10&CONV, ReLU, MP, FC&CE&Inference time(s)=15.05\\

\cite{xie2015beyond}&Breast&H\&E&40x&Cell detection&32&--&TCGA&$49\times49$&--&--&CNN&8&CONV, ReLU, MP, FC&WCE&P=0.91, R=0.91, F1=0.91\\

\cite{xie2015deep}&Neurones&Ki-67&20x&Nuclei detection&44&--&--&$39\times39$&--&--&CNN&8&CONV, ReLU, MP, FC, DO&WL2+L1&P=0.85, R=0.79, F1=0.81\\

\cite{sirinukunwattana2015spatially}&Colon Breast&H\&E&20x&Nuclei detection&30&&ICPR2014&$27\times27$&L*ab conversion&Rotation Morroring&CNN&6&CONV, ReLU, MP, FC&WCE&P=0.71/0.73, R=0.85/0.78, F1=0.77/0.75\\

\cite{sirinukunwattana2016locality}&Colon&H\&E&20x&Nuclei detection &100&20K&CRCHistoNuclei&$27\times27$&CD&Affine&CNN& 8 &CONV, ReLU, MP, FC&WCE&P=0.75, R=0.83, F1=0.79\\

\cite{kashif2016handcrafted}&--&H\&E&--&Nuclei detection&15&--&--&$27\times27$&--&--&CNN&8&CONV, ReLU, MP, FC&CE&P=0.77, R=0.72, F1=0.75\\

\cite{xie2018microscopy}&Blood Retina&H\&E&--&Cell counting&2&7000&--&$100\times100$&--&Synthetic IF images&FCN&7&CONV, ReLU, MP, FC&L2&True/Estimated count=705/696\\

\cite{chen2016automated}&Breast&H\&E&400x&Mitosis detection&50&300&--&2K$\times$2K&--&--&DeepLab&16&CONV, ReLU, MP, FC&CE&F1=0.79\\

\cite{veta2016cutting}&Breast&H\&E&40x&Nuclei area&39&4264&--&$96\times96$&--&--&CNN&10&CONV, ReLU, MP, FC&CE&Bland-Altman Bias=-2.98\\

\cite{song2015accurate}&Cervex&H\&E&40x&Cell segm.&50&8590&--&$32\times32$&YUV color conversion&Multi-scale &CNN&5&CONV, Tanh, MP, FC&CE&DICE=0.95\\

\cite{janowczyk2018resolution}&Breast&H\&E&40x&Nuclei segm.&137&12K&--&$32\times32$&--&Multi-scale,  Boosting &AlexNet &6&CONV, ReLU, MP, FC, DO&CE&F1=0.82, TPV=0.81, PPV=0.88\\

\cite{song2017accurate}&Cervex&H\&E&--&Cell segm.&8&20-60&ISBI2015-CELL&$32\times32$&YUV color conversion&Rotations&CNN&3&CONV, ReLU, MP, FC&CE&DICE=0.89\\

\cite{xing2016automatic}&Brain Breast&H\&E&--&Nuclei segm.&30/35&600K&--&$55\times55$&YUV color conversion&Rotations&CNN&6&CONV, ReLU, MP, FC&CE&F1=0.77/0.78\\

\cite{de2018automatic}&Kidney&H\&E&20x&Kidney structures segm.&15&3518&--&$100\times100$&&Elastic&U-Net&22&CONV, ReLU, MP, FC&CE&ConfMat.\\

\cite{sirinukunwattana2017gland}&Colon&H\&E&40x&Gland segm.&165&--&GLaS&--&SNlization&Elastic&FCN&16&CONV, ReLU, MP, UP&CE&DICE=0.78, Hausdorff=160.3, F1=0.72 \\

\cite{chen2017dcan}&Colon&H\&E&40x&Gland segm.&165&--&GLaS &$480\times480$&--&Affine \& elastic &FCN&8&CONV, ReLU, MP, UP &Multi-loss&DICE=0.78, Hausdorff=160.3, F1=0.72\\

\cite{raza2017mimo}&Pancreas&IF&40x&Cell segm.&--&11K&--&$252\times252$&&Gaussian noise lens distortion flip and rotate&ResNet&50&Residual blocks&WCE&DICE=0.8, ObjectDICE=0.84, Hausdorff=27.5, F1=0.72\\
%\multicolumn{17}{|c|}{LARGE ORGANS}\\

\cite{cruz2013deep}&Skin&H\&E&10x&Basal Cell Carcinoma detection&1417&--&BCC&$8\times8$&YUV conversion&--&SSAE&&CONV, ReLU,AP,FC&MSE&ACC=0.91\\

\cite{cruz2014automatic}&Breast&H\&E&40x&Ivasive Ductal Carcinoma detection&162&--&--&$100\times100$&--&Bootstrap&CNN&3&CONV, ReLU, MP&CE&ACC=0.84\\

\cite{litjens2016deep}&Prostate Breast &H\&E&--&Cancer detection &254/271&--&--&$128\times128$&--&--&CNN&6&CONV, ReLU, MP&CE&AUC=0.99/0.88\\

\cite{cruz2017accurate}&Breast&H\&E&40x&Cancer detection &584&--&TCGA&$101\times101$&--&--&CNN&3&CONV, ReLU, MP&CE&DICE=0.76\\

\cite{bejnordi2017diagnostic}&Breast&H\&E&40x-20x&Metastasis dection&399&--&Camelyon17&$296\times296$&--&Affine&Inception&&Inception blocks&CE&AUC=0.99\\

\cite{golden2017deep}&Breast&H\&E&--&Metastasis detection&399&--&Camelyon17&$296\times296$&--&Affine&Inception&&Inception blocks&CE&AUC=0.99\\

\cite{kong2017cancer}&Breast&H\&E&40x&Metastasis detection&399&--&Camelyon17&$256\times256$&--&Affine&ResNet&101&Residual blocks&WCE&FROC=0.75\\

\cite{bauer2016multi}& Prostate Kidneys&H\&E&--&Nuclei classif. &8/6&--&--&--&--&Rotation&Resnet&50&Residual blocks&CE&F1=0.99\\

\cite{phan2016transfer}&Hep2&IF&--&Cell classif.&28&1457&ICPR2012&$224\times224$&--&--&AlexNet&8&CONV, ReLU, MP, FC&CE&ACC=0.77\\

\cite{qaiser2018her}&Breast&IHC&40x&Image scoring&172&--&Her2&&--&--&AlexNet Inception&8/16&CONV, ReLU, UP&CE&Agreement Points=382.5\\

\cite{yao2016imaging}&Lung&H\&E&--&Cell classif.&257&--&TCIA-NLST&$40\times40$&--&--&FCN&3&CONV, ReLU, MP&CE&Kaplan-Meier curves\\

\cite{xu2016deep}&Epithelium&IHC H\&E&20x&Tissue classif.&157/27&--&Released&$80\times80$&--&--&FCN&5&CONV, ReLU, MP&CE&ACC=100\\

\cite{liu2017detecting}&Breast&H\&E&40x&Tumor detection and segm.&270&--&Camelyon16&$299\times299$&--&--&Inception&16&CONV, ReLU, MP&CE&FROC\\

\cite{nazeri2018two}&Breast&H\&E&--&Tissue classif.&400&--&ICIAR18&$512\times512$&--&Rotations&CNN&15/6&CONV, ReLU, MP&CE&AUC=0.98\\

\cite{ertosun2015automated}&Brain&H\&E&20x&Malignant gliomas grading&22&7066&TCGA&$256\times256$&--&--&CNN&6/12&CONV, ReLU, MP&CE&ConfMat\\

\cite{nayak2013classification}&Brain Kidney&H\&E&40x&Morphometric signatures classif.&2500/ 1400&--&TCGA-KIRC&$100\times100$&--&Rotation, flip&SAE&2&FC&MSE+L2&ConfMat\\

\cite{bejnordi2017deep}&Breast&H\&E&20x&Cancer classif.&646&--&--&$224\times224$&--&Affine&VGG&16&CONV, ReLU, MP,UP&WCE&AUC=0.92\\

\cite{wang2016deep}&Breast&H\&E&40x&Metastatic cancer detection and classif.&400&--&Camelyon16&$256\times256$&--&Rotations, flips, crops&GoogleNet&27&CONV, ReLU, MP, BN&CE&AUC=0.93\\

\cite{graham2018classification}&lung&H\&E&20X&Cancer grading&64&65K&MICCAI-CPM17&$224\times224$&--&--&ResNet&32&Residual blocks&CE&ACC=0.81\\

\cite{ren2018differentiation}&Prostate&H\&E&40x&Gleason grading &270&--&TCGA GDC&$256\times256$&--&--&CNN LSTM& -- &LSTM units&CE&Hazard=206\\

\cite{hou2016patch}&Brain lung&H\&E&--&Lung cancer classif.&1064&1.1M&TCGA-NSCLC&$500\times500$&CD&Rotation, mirror, color jitter&CNN&8&CONV, ReLU, MP, LRN, DO&CE&mAP=0.85\\

\cite{yang2017suggestive}&Breast&H\&E&&Lymph node segm.&85&--&GLaS&--&--&--&FCN&50&Residual blocks&CE& Hausdorff=96.9, DICE=0.86, F1=0.86\\

\cite{agarwalla2017representation}&Breast&H\&E&40x&Cancer segm.&270&12M&Camelyon16&$224\times224$&Background removal&--&CNN-LSTM&10&LSTM units&CE&F1=0.83, R=0.83, P=0.81\\

\cite{wang2017adversarial}&Cervix&H\&E&40x&Membrane segm.&200&--&--&$500\times500$&--&Rotations flips&VGG&21&CONV, ReLU, MP, UP&ADV&F1=0.62, P=0.61, R=0.64\\

\cite{qaiser2017tumor}&Colon&H\&E&20x&Tumor segm.&50&50K&--&$256\times256$&--&--&CNN&10&CONV, ELU, MP, FC&CE&F1=0.90, P=0.88, R=0.92\\

\cite{jia2017constrained}&Colon&H\&E&--&Cancer segm.&930&--&--&$64\times64$&--&--&VGG&16&CONV, ReLU, MP, UP&MIL CE&F1=0.83\\

\cite{bychkov2016deep}&Colon&H\&E&--&Survival &180&--&--&$585\times585$&--&--&CNN&9&CONV, ReLU, MP, FC&CE+L2&Hazard=2.08, CI=0.95, AUC=0.66\\

\cite{bauer2016multi}&Prostate Kidneys&H\&E&--&Nuclei classif.&6/8&826/1278&--&$78\times78$&Grayscale conversion&Affine&ResNet&34&Residual blocks&CE&F1=0.82/0.99\\

\cite{zhu2017wsisa}&Lung Brain&H\&E&20x&Survival &1104 485 255&67K 70K 60K&NLST/TCIA&$512\times512$&Background removal&--&CNN&6&CONV, ReLU, MP, FC&Cox regr.&C-index=0.7/0.63/0.60\\

\cite{zhu2016deep}&Lung&H\&E&--&Survival &450&--&NLST&$339\times339$&--&--&CNN&6&CONV, ReLU, MP, FC&Cox regr.&C-index=0.63\\

\cite{yao2017deep}&Lung Brain&H\&E Omics&--&Survival &106/126&--&NLST&$1024\times1024$&--&--&CNN&7&CONV, ReLU, MP, FC&Cox regr.&C-index=0.63/0.64\\

\cite{mobadersany2018predicting}&Brain&H\&E Omics&20x&Survival &769&--&TCGA&$256\times256$&SN&Mirror, Color jitter&VGG&19&CONV, ReLU, MP, LN, FC&Cox regr.&C-index=0.75\\

%\multicolumn{17}{|c|}{OTHER TASKS}\\
\cite{saltz2018spatial}&Breast&H\&E&20x&Computational staining&5455&--&TCGA&$100\times100$&Background removal&Boosting, color jitter&CNN-AE&18&CONV, ReLU, MP, FC&MSE+CE&AUC=0.95\\

\cite{janowczyk2017stain}&Breast Colon&H\&E&40x&SN&25&200&--&$1000\times1000$&--&--&SSAE&--&FC&MSE&DICE=0.76\\

\cite{zanjani2018stain}&Breast Liver&H\&E&--&SN&125&625&--&$299\times299$&--&--&InfoGAN&--&--&ADV+R&NMI=0.036\\

\cite{lafarge2018inferring}&Breast&H\&E&--&Data augm.&238&--&AMIDA2013&$24\times24$&CD&--&CNN&6&CONV, lReLU, MP, FC&CE &Qualitative\\

\cite{moeskops2017domain}&Breast&H\&E&--&Domain adapt.&73&1552&TUPAC&$63\times63$&--&Color jitter&CNN&6&CONV, ReLU, MP, BN, FC&Multi-loss CE&F1=0.62\\

\cite{senaras2018creating}&Breast&IHC&40x&Data synthesis&32&--&--&$512\times512$&--&Color jitter, rotations, flips&cGAN&--&CONV, lReLU, MP, FC&ADV&Experts Fooled ACC=0.47\\

\cite{bayramoglu2017towards}&Lung&None&40x&Digital staining&--&2838&--&$64\times64$&PCA &--&cGAN&--&CONV, lReLU, MP, FC&ADV&SSIM=0.39, MSE=2.44\\

\cite{cho2017neural}&Breast&H\&E&40x&Stain transfer&400&180K&Camelyon16&--&Grayscale conversion&--&cGAN&39&CONV MP FC UP ReLU&ADV+R&AUC=0.91, P=0.84, R=0.85, SP=0.84\\

\cite{burlingame2018shift}&Pancreas&H\&E, IF&10x-20x&Digital staining&--&20K&--&$256\times256$&Registration&Color jitter, rotations, flips&cGAN&--&CONV, ReLU, MP, UP, FC&ADV+R&DICE=0.9, PSNR=31.5, SSIM=0.9\\

\cite{schuler2017context}&Lung&H\&E&--&Stain transfer&5&--&--&$192\times192$&--&--&CNN&--&CONV, ReLU, MP, FC&MSE&Qualitative\\

\cite{tomczak2016improving}&Bone marrow&H\&E&--&Data synthesis&16&--&Enjoypath&$28\times28$&Grayscale conversion&--&VAE&--&FC&Lower bound on log-likelihood&NATS=1398.3\\

\cite{korbar2017looking}&Colon&H\&E&--&Feature visualization&176&--&--&$224\times224$&--&--&ResNet&152&Residual blocks&CE &IOU=0.55\\

\end{longtable}
}
\end{landscape}

\bibliographystyle{plain}
\bibliography{main}

\end{document}